\documentclass[10pt,journal,compsoc]{IEEEtran}

\ifCLASSOPTIONcompsoc

  \usepackage[nocompress]{cite}
\else
  \usepackage{cite}
\fi

\ifCLASSINFOpdf

\else

\fi

\usepackage{graphicx}
\usepackage{amsmath}
\usepackage{amssymb}
\usepackage{booktabs}
\usepackage{multirow}

\usepackage{bbm}
\usepackage{adjustbox}

\usepackage{amsmath,amsfonts,bm}

\def\eqref#1{equation~\ref{#1}}

\def\1{\bm{1}}

\DeclareMathAlphabet{\mathsfit}{\encodingdefault}{\sfdefault}{m}{sl}
\SetMathAlphabet{\mathsfit}{bold}{\encodingdefault}{\sfdefault}{bx}{n}

\usepackage[usenames,dvipsnames]{xcolor}
\usepackage{xcolor}
\usepackage{pifont}
\usepackage{algorithm}
\usepackage{listings}
\usepackage{comment}
\usepackage{makecell}
\usepackage{lipsum}  

\usepackage{resizegather}

\renewcommand{\paragraph}[1]{\vspace{1.25mm}\noindent\textbf{#1}}

\usepackage[pagebackref,breaklinks,colorlinks]{hyperref}

\usepackage[capitalize]{cleveref}
\crefname{section}{Sec.}{Secs.}
\Crefname{section}{Section}{Sections}
\Crefname{table}{Table}{Tables}
\crefname{table}{Tab.}{Tabs.}

\begin{document}

\title{A Survey on Diffusion Model for Generating Graph, Molecules,  Materials and Beyond}

\title{A Survey on Diffusion Model in Graph for Generating Molecules,  Materials and Beyond}

\title{A Survey on Graph Diffusion Model for Molecule Generation and Beyond}

\title{Graph Diffusion Model: A Survey on Molecule and Protein Generation in Generative AI}

\title{Graph Diffusion Model Survey: Molecule and Protein Generation in Generative AI}

\title{A Survey on Graph Diffusion Models: AIGC in Science for Molecule, Protein and Material}

\title{A Survey on Graph Diffusion Models: Generative AI in Science for Molecule, Protein and Material}

\author{Mengchun~Zhang*, Maryam~Qamar*\thanks{* indicates equal contribution}, Taegoo Kang, Yuna~Jung, Chenshuang~Zhang, Sung-Ho Bae$\dagger$, Chaoning Zhang$\dagger$ \thanks{$\dagger$ Correpondence author(s)}%
\IEEEcompsocitemizethanks{
\IEEEcompsocthanksitem Mengchun Zhang and Chenshuang Zhang are with KAIST (South Korea). Maryam Qamar, Taegoo Kang, Yuna Jung, Sung-Ho Bae, Chaoning Zhang are with Kyung Hee University (South Korea). \protect\\

\IEEEcompsocthanksitem Chaoning Zhang (chaoningzhang1990@gmail.com) and Sung-Ho (shbae@khu.ac.kr) are correspondence authors. \protect\\ 

}

}

\markboth{Journal of \LaTeX\ Class Files,~Vol.~14, No.~8, August~2015}
{Zhang \MakeLowercase{\textit{et al.}}: A Survey on Graph Diffusion Models}

\IEEEtitleabstractindextext{%
\begin{abstract}

Diffusion models have become a new SOTA generative modeling method in various fields, for which there are multiple survey works that provide an overall survey. With the number of articles on diffusion models increasing exponentially in the past few years, there is an increasing need for survey of diffusion models on specific fields. In this work, we are committed to conducting a survey on the graph diffusion models. Even though our focus is to cover the progress of diffusion models in graphs, we first briefly summarize how other generative modeling methods are used for graphs. After that, we introduce the mechanism of diffusion models in various forms, which facilitates the discussion on the graph diffusion models. The applications of graph diffusion models mainly fall into the catogory of AI-generated content (AIGC) in science, for which we mainly focus on how graph diffusion models are utilized for generating molecules and proteins but also cover other cases, including materials design. Moreover, we discuss the issue of evaluating diffusion models in the graph domain and the existing challenges.

\end{abstract}

}

\maketitle

\IEEEdisplaynontitleabstractindextext

\IEEEpeerreviewmaketitle

\IEEEraisesectionheading{\section{Introduction}\label{sec:introduction}}

\IEEEPARstart{G}{raphs} are ubiquitous in various scenarios. Due to the great expressive power of graphs, a growing number of studies have merged using machine learning for graph analysis and generation in recent years~\cite{sun2009concise,mikolov2013efficient,kipf2016semi}. Graphs are broadly implemented in a considerable number of research domains, including social networks in social sciences~\cite{hamilton2017inductive,kipf2016semi}, molecular structures in chemistry and materials~\cite{chen2019graph}, knowledge bases~\cite{hamaguchi2017knowledge} and protein interaction networks in biochemistry~\cite{bacciu2020gentle}. Graphs have an irregular structure with varying numbers of nodes and edges, which make it difficult to apply standard convolution or pooling operations~\cite{duvenaud2015convolutional,bruna2013spectral}. Moreover, graphs can be large in scale, requiring specialized techniques to process and analyze them efficiently~\cite{kyrola2012graphchi,malewicz2010pregel,gonzalez2012powergraph}. Moreover, graphs exhibit diverse and heterogeneous properties, making it challenging to develop models that can generalize well across different types of graphs~\cite{hamilton2017representation}. These challenges have motivated researchers to develop new approaches and architectures for deep learning on graph data, such as graph convolutional networks and graph attention networks, that can effectively capture the structure and properties of graphs~\cite{zhang2020deep,hamilton2017representation,wang2019heterogeneous,chiang2019cluster,zhu2020beyond}. 

Representing molecules and proteins as graphs enables researchers to harness the power of graph neural networks to learn their structural and functional properties, which can then be employed to create novel molecules and proteins with expected characteristics~\cite{wu2018moleculenet,fout2017protein}. Molecular and protein generation tasks involve employing computational approaches to design novel molecules and proteins or optimize existing ones, which are more efficient and cost-effective than conventional experimental techniques~\cite{ingraham2019generative,popova2018deep}. Researchers can create vast libraries of potential molecules and proteins by leveraging graph-based deep-learning techniques. This method can considerably accelerate the finding of novel drugs and materials and potentially revolutionize the field of molecular design~\cite{bjerrum2017molecular,jing2018deep}.

Diffusion models have become an essential tool in graph-based deep learning, providing a method to model the diffusion of information across the nodes and edges of a graph. With applications varying from representation learning to graph generation and dynamic graph processing, diffusion models have demonstrated versatility in addressing various challenges in graph-based deep learning. Furthermore, diffusion models have been shown to improve the scalability and efficiency of graph neural networks, specifically for large-scale graphs~\cite{li2015gated,ingraham2019generative}. They also provide a way to incorporate global information into the learning process, leading to more precise predictions and better performance on downstream tasks.

There are extensive works covering the graph generation techniques based on autoregressive, VAE, and GAN models ~\cite{you2018graphrnn, popova2019molecularrnn, simonovsky2018graphvae, de2018molgan}. However, the literature lacks in a comprehensive review of this task regrading the emerging and promising diffusion models. In this survey, we strive to remedy this knowledge gap. We approach this by spelling out a brief summary of the aforementioned methods from graph generation perspective along with the challenges they face in section \ref{dgg}. The survey then progresses to the main topic of diffusion based approaches where section \ref{bgdm} introduces the diffusion models succinctly and then we move on to their deep discussion for graph generation in section \ref{gdm} and \ref{app}, both from structural and application specific point of view. The survey then proceeds to explain the currently popular evaluation datasets and metrics in section \ref{eva}, concluding with the challenges and research questions still open to the research community in section \ref{ch}. By authoring this in-depth survey, we hope to provide a comprehensive resource for researchers interested in exploring the potential of diffusion models for graph generation and related tasks.

\textbf{Related survey works.} Multiple works~\cite{yang2022diffusion,ulhaq2022efficient,cao2022survey} survey the progress of diffusion models in all fields. There are also multiple field-specific survey works on diffusion models, including text-to-image diffusion models~\cite{zhang2023text}, audio diffusion models~\cite{zhang2023audio}. Complementary to them, this work conducts a survey on graph diffusion models. From the perspective of AI-generated content (AIGC), this survey is also related to generative AI (see~\cite{zhang2023complete} for a survey) and ChatGPT (see ~\cite{zhang2023ChatGPT} for a survey). Conceptually, diffusion models are also highly related to masked autoencoders (see~\cite{zhang2022survey} for a survey), which also perform denoising.

\section{Deep Graph Generation}
\label{dgg}

Generative models such as the Erdos-Renyi~\cite{erdHos1960evolution}, Watts-Strogatz~\cite{watts1998collective}, Barabasi-Albert~\cite{barabasi1999emergence}, and R-MAT~\cite{chakrabarti2004r} observe manually specific significant properties of actual graphs and construct algorithms accordingly. While these models may successfully capture some features of actual graphs, they lack the flexibility to model different graph properties~\cite{leskovec2010kronecker,leskovec2010kronecker,chiang2019cluster,hamilton2017inductive}. Furthermore, generative models may struggle to generate graphs that indicate observed connectivity patterns or behaviors~\cite{bojchevski2018netgan}. As a result, there is a need to develop more sophisticated generative models that can integrate a more comprehensive scope of graph properties and yield more realistic results~\cite{yang2016revisiting}.

\begin{figure*}[t]
    \centering
    \includegraphics[width=0.90\linewidth]{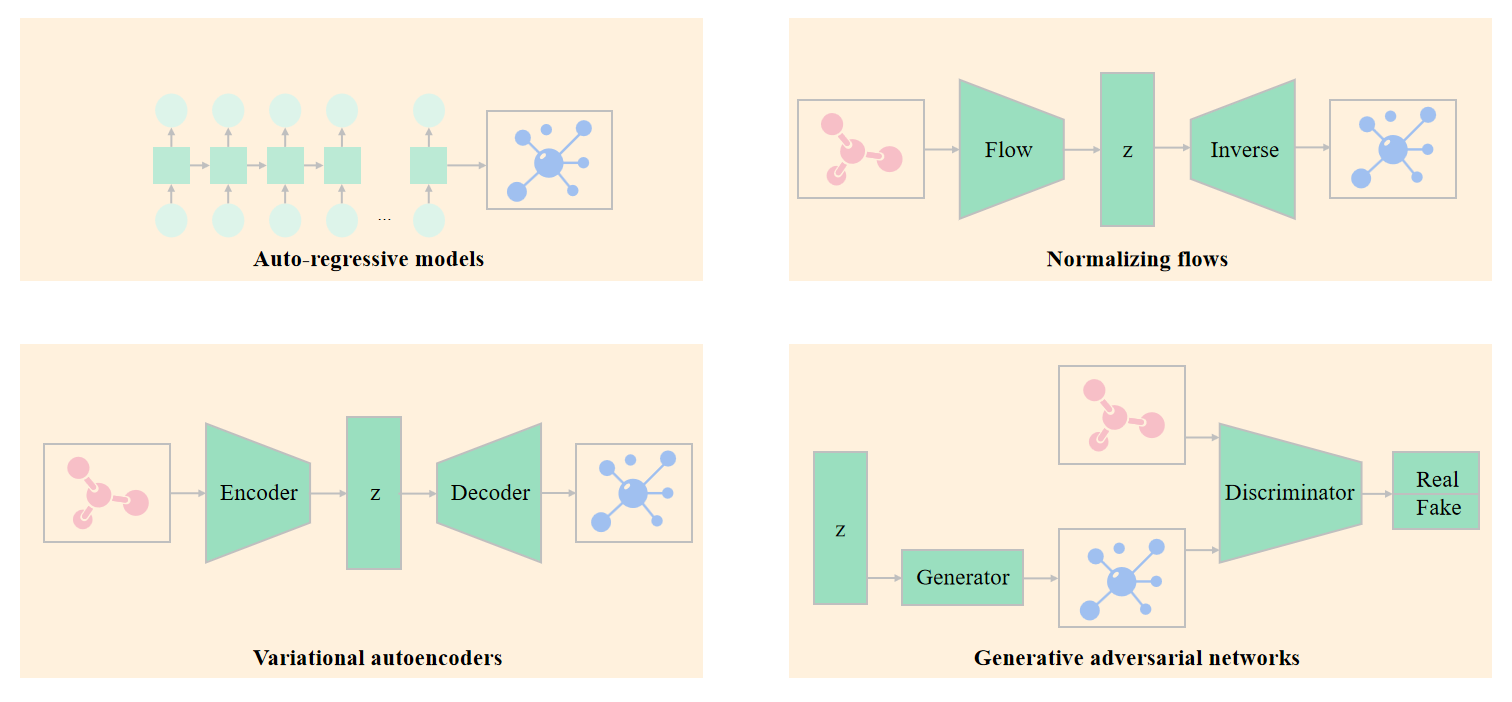}
    \caption{Graph generation models for deep graph generation}
    \label{figure_1}
\end{figure*}

In this section, we briefly review the deep graph generation models according to the algorithmic taxonomy. Regardless of data type, there are four popular generative modeling methods, namely, Auto-Regressive Models~\cite{vaswani2017attention}, Variational Auto-Encoders~\cite{kingma2013auto}, Normalizing Flows~\cite{dinh2016density}, and Generative Adversarial Networks~\cite{goodfellow2020generative}. Here, we summarize the unified framework for deep graph generation models as shown in Figure~\ref{figure_1}.

\subsection{Auto-Regressive Model}
Auto-Regressive model(AR) factorizes the generation process into a sequence of steps, which predicts the next output based on the previous output in the sequence~\cite{vaswani2017attention}. The most common objective function used in Auto-Regressive generation models is the maximum likelihood estimation (MLE), which leads the model to generate sequences that are similar to those in the training data~\cite{sutskever2014sequence,radford2017learning,reed2017parallel}. In the field of graph generation, GraphRNN~\cite{you2018graphrnn} uses two Recurrent Neural Networks (RNNs) in order to learn the distribution of numerous representations of graphs that vary with node orderings. Each RNN learns to generate the sequence of nodes and edges for graph structures including newly added nodes, so that it can capture the complex structural dependencies of the graph~\cite{gilmer2017neural,ying2018graph}. Inspired by GraphRNN~\cite{you2018graphrnn}, MolecularRNN~\cite{popova2019molecularrnn} adopts this framework to generate molecular structure. In addition, they use policy gradient optimization algorithm to produce molecules that satisfy certain conditions. GraphAF~\cite{shi2020graphaf} combines AR and Normalizing Flows, defining the sampling process in a sequential way using flow-based graph neural network. Similar to \cite{popova2019molecularrnn}, they use reinforcement learning to generate valid and conditional molecules. Despite AR showing remarkable performance in graph generation, they suffer from designing permutation-invariant graph distribution since they treat graph as a sequence~\cite{you2018graphrnn,bui2018neural,you2018graph}.\\

\subsection{Variational Auto Encoder}
Variational Auto Encoder(VAEs) consists of an encoder that maps the input data into a low-dimensional latent space and a decoder that maps latent representation back to the original data space~\cite{kingma2013auto,doersch2016tutorial}. VAEs are optimized by minimizing the reconstruction error between the generated output and the input data, while regularizing the latent space to follow a prior probability distribution. VAE-based methods are categorized into two ways: one-shot~\cite{simonovsky2018graphvae,ma2018constrained,grover2019graphite} and sequential generation~\cite{liu2018constrained,jin2018junction}. For the one-shot type, VGAE~\cite{kipf2016variational} generates the edge probability based on the relationships between nodes’ embedding. In GraphVAE~\cite{simonovsky2018graphvae}, the decoder outputs a probabilistic fully connected graph in one-step. To find the best graph, they use approximate graph matching between nodes of graphs based on the similarities of their node pairs. When generating a graph, there is often a constraint demand that only specific combinations of the nodes and edges are valid in context~\cite{you2018graph,simonovsky2018graphvae,liu2018constrained}. RGVAE~\cite{ma2018constrained} generates a semantically valid graph by adding a penalty term that regularizes the output distribution of the decoder to the objective function of VAEs. Graphite~\cite{grover2019graphite} repeats the process that constructs an intermediate weighted graph and iteratively refines the reconstructed feature matrix via message passing. The final distribution of graph is obtained via the inner product of feature matrix. The other category is sequential generation method. CGVAE~\cite{liu2018constrained}, node-by-node generator, repeats the process that chooses which node is connected to the current node and selects and labels edges between nodes to generate a graph. By contrast, JT-VAE~\cite{jin2018junction}, a motif-sequence-based generator, generates molecular graphs by a procedure that generates a tree-structured scaffold over chemical substructures first and then combines them into a molecule. But Autoencoder-bassed graph generation methods have limitations in that they are not scalable to be applied on large graphs.

\subsection{Normalizing Flows}
Normalizing Flow applies a sequence of invertible transformations to a simple probability distribution to model more complex probability distributions using encoder~\cite{rezende2015variational,dinh2016density,papamakarios2017masked,ho2019flow++}. Its decoder is essentially an inverse function of the encoder. GraphNVP~\cite{madhawa2019graphnvp} first adopts normalizing flows to molecular graph generation. To address the lack of validity constraint in the one-shot generation model, they propose a two-step generation process which is consisted of generating a graph structure and node features according to the generated structure. GNF~\cite{liu2019graph} proposes a new graph neural network reversible for prediction and generation. They use graph auto-encoder that maps discrete graph structure to continuous graph embedding. Then GNF learns the distribution of graph embedding, which is obtained by a learned graph autoencoder. MoFlow~\cite{zang2020moflow} uses two invertible flow-based networks, one is for generating bonds and another is for generating atoms with given bonds. They use post-hoc method to correct the validity of the molecule. Unlike most previous methods, which map discrete graph structure to continuous latent variables, GraphDF~\cite{luo2021graphdf} uses discrete latent variables using invertible transform. One of the critical limitations of Normalizing Flows is that there are constraints on generating valid graphs due to its one-shot nature~\cite{liu2019graph}. \\

\subsection{Generative Adversarial Networks}
Generative Adversarial Networks (GANs) are composed of two main components: a generator network that generates synthetic data for deceiving the discriminator, and a discriminator network that assesses the authenticity of the generated data~\cite{creswell2018generative,mirza2014conditional,radford2015unsupervised,salimans2016improved}. GAN-based methods utilize the property of discriminator to generate a graph to have desired properties. MolGAN~\cite{de2018molgan} is the first address to generate a small molecular graph using GANs. The generator samples from the prior distribution and outputs the entire probabilistic graph at once. The discriminator distinguishes whether the molecule comes from the generator or the dataset. And they further utilize reward network for RL-based optimization towards desired chemical properties. There are also graph-to-graph translation methods that generate graphs with better properties from source graphs, such as VJTNN~\cite{jin2018learning} and Mole-CycleGAN~\cite{maziarka2020mol}. Unlike the above methods that generate graphs, there are methods of generating graphs via random walks that are invariant under node reordering. NetGAN~\cite{bojchevski2018netgan} generates random walks that are plausible in the real graph, and then they are used to produce an adjacency matrix of a newly generated graph. But GAN-based methods are challenging for graph generation tasks because of the need for likelihood-based optimization.

\section{Background on diffusion models}
\label{bgdm}

Diffusion models are a class of generative models that gradually introduce noise into data until it conforms to a prior distribution~\cite{ho2020denoising}. The models then learn to reverse this process to generate viable samples. By leveraging the power of diffusion models, researchers can create generative models that can accurately capture the underlying structure of complex datasets and produce high-quality, realistic samples.

\begin{figure}[!t]
    \centerline{\includegraphics[width=\columnwidth]{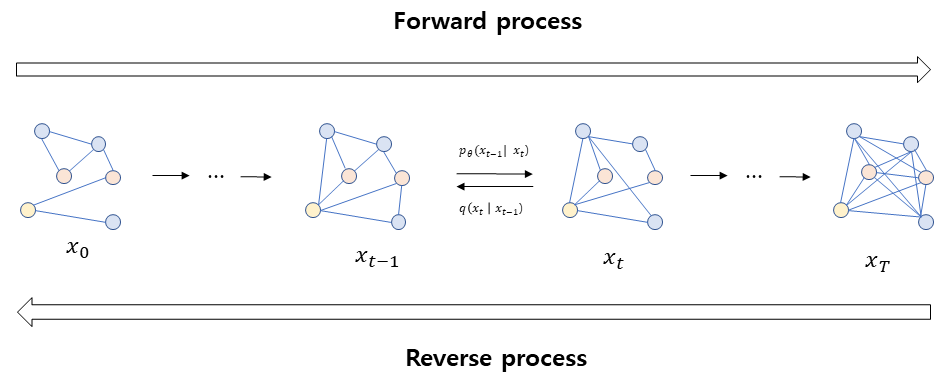}}
    \caption{Illustration of diffusion models for deep graph generation.}
    \label{figure_2} 
\end{figure}

There are three sub-types: denoising diffusion probabilistic models (DDPMs) ~\cite{vignac2022digress, haefeli2022diffusion, anand2022protein, trippe2022diffusion, luo2022antigen}, score-based generative models (SGMs) ~\cite{niu2020permutation, chen2022nvdiff}, and stochastic differential equations (SDEs) ~\cite{huang2022graphgdp, jo2022score, luo2022fast}, which differ in how they implement the forward and backward diffusion pass. Figure~\ref{figure_2} presents a generic framework that is based on three categories.

\subsection{Denoising Diffusion Probabilistic Models(DDPM)}

A denoising diffusion probabilistic model (DDPM) is a generative model that can generate novel data samples from a given data distribution that employs two Markov chains~\cite{ho2020denoising}.

\textbf{Forward process.}
First, the original data is transformed into a simpler prior distribution by gradually adding noise over a fixed number of diffusion steps T according to a variance schedule $\beta$. This forward pass aims to transform the input data into a distribution that can be easily sampled using standard techniques such as sampling from a Gaussian distribution. Given a data distribution $\mathbf{x}_0 \sim q(\mathbf{x}_0)$, the forward process generates $\mathbf{x}_T$ with transition kernel ${q(x_t|x_{t-1})}$ and can be defined as follows~\cite{ho2020denoising}:

\begin{align}
q(x_{1:T} | x_0) := \prod_{t=1}^T q( x_t | x_{t-1} ), \label{eq:forwardprocess_1}
\end{align}
\begin{align}
q(x_t|x_{t-1}) := \mathcal{N}(x_t;\sqrt{1-\beta_t} x_{t-1},\beta_t I) \label{eq:forwardprocess_2}
\end{align}

where $\beta_t$ is a hyper-parameter. To simplify the discussion, we focus on the use of Gaussian noise as the transition kernels denoted as $\mathcal{N}$ in \ref{eq:forwardprocess_2}. When $\alpha_t := 1 - \beta_t$ and $\bar{\alpha}_t := \prod_{s=0}^{t} \alpha_s$:
\begin{align}
q(x_t|x_{0}) := \mathcal{N}(x_t;\sqrt{\bar{\alpha}_t}x_{0},(1 - \bar{\alpha}_t) I) \label{eq:forwardprocess_3}
\end{align}

\textbf{Reverse process.} To generate new samples from the data distribution, the reverse process is performed. The reverse pass makes use of a neural network trained to predict the noise that was added at each step in the forward pass. This pass removes the noise at each step in reverse order until the original data is recovered. To generate $p_\theta(x_0)$ that follows the true data distribution $q(x_0)$, we start with $p_\theta(T)$ and optimize the model with the following objective, as described in~\cite{ho2020denoising}:
\begin{equation}
\label{eq:reverseprocess_1}
p_\theta( \mathbf{x}_{t-1} | \mathbf{x}_{t} ) := \mathcal{N}(\mathbf{x}_{t-1};\mu_\theta(x_t,t),\sigma^2_t \mathbf{I})
\end{equation}

As stated in~\cite{yang2022diffusionsurvey}, the model's optimization objective can be expressed as:
\begin{align}
     E_{t \sim \mathcal{U}( 1,T ), \mathbf x_0 \sim q(\mathbf x_0), \epsilon \sim \mathcal{N}(\mathbf{0},\mathbf{I})}{ \lambda(t)  \left\| \epsilon - \epsilon_\theta(\mathbf{x}_t, t) \right\|^2} \label{eq:loss}
\end{align}

DiGress~\cite{vignac2022digress} and ~\cite{haefeli2022diffusion} adopt DDPM technique for graph diffusion though they add discrete noise instead of continuous Gaussian.

\subsection{Score-Based Generative Models(SGMs)}
Score-based generative models are a class of probabilistic models that employ a score function, also referred to as an energy function or an unnormalized probability density function, to illustrate the probability distribution of the data~\cite{song2020improved,song2020sliced}.

The score function of a probability density function $p(x)$ is defined as the gradient of the logarithm of the probability density, $\nabla_x \log p(x)$. A noise-conditional score network is a type of deep neural network that is trained to estimate the score function $\nabla_{x_t} \log q(x_t)$, which represents the gradient of the logarithm of the conditional probability density function $q(x_t)$ of the data given the noise. The network is denoted as $s_\theta(x, t)$, where $\theta$ represents the learnable parameters of the network and $x$ means the input data at time $t$. The network is generally trained using maximum likelihood estimation or a variant thereof, such as noise-contrastive estimation or denoising score matching.

Employing denoising score matching and similar notations, as represented in \ref{eq:loss}, the training objective for the score network is given by:

\begin{align}
     \mathbb{E}_{t \sim \mathcal{U}[[1,T]] ), \mathbf x_0 \sim q(\mathbf x_0), \epsilon \sim \mathcal{N}(\mathbf{0},\mathbf{I})}[ \lambda(t)  \left\| \mathbf{\epsilon} + \sigma_t s_\theta(\mathbf{x}_t, t) \right\|^2] \label{eq:loss_2}
\end{align}

The pioneer score-based graph generation model, EDP-GNN~\cite{niu2020permutation}, leverages GIN~\cite{xu2018powerful} and makes use of annealed Langevin dynamic sampling to diffuse the edges. In contrast, NVDiff~\cite{chen2022nvdiff}, utilizes  an attention-based score generative model to exclusively diffuse the latent representation of node vectors.

\subsection{Stochastic Differential Equations (SDEs)}
Stochastic Differential Equations (SDEs) are a class of mathematical models that characterize the development of a system over time under the effect of random noise. A Score-based Stochastic Differential Equation (Score SDE) is a type of SDE where the drift term is defined as the negative gradient of a score function, and the diffusion term is a function of time~\cite{song2020score}.

\textbf{Forward SDE.}
In mathematics, a forward SDE is a continuous-time dynamical system that describes the evolution of a state variable over time, where both deterministic and stochastic forces govern the evolution. The forward SDE can be defined as follows~\cite{song2020score}:
\begin{equation}
dx={f(x,t)}dt+g(t)dw
\end{equation}

\textbf{Reverse SDE.}
A reverse stochastic differential equation (SDE) is a continuous-time dynamical system that progresses backward. It is generally employed to calculate the score function for the forward SDE, which can subsequently be operated to produce samples from the conditional distribution. The reverse SDE can be defined as follows~\cite{song2020score}
\begin{equation}
dx = [f(x,t)-g(t)^2 \nabla_x \log p_t(x)]dt + g(t)d\bar{w}
\end{equation}
The score function is estimated by parameterizing a score model $s_{\theta}(x_t, t)$, which involves generalizing the score matching objective in \ref{eq:loss_2} to continuous time as follows~\cite{yang2022diffusion}:
\begin{align}
     \mathbb{E}_{t \sim \mathcal{U}[0,T] ), \mathbf x_t \sim q(\mathbf x_t | \mathbf x_0)}[ \lambda(t) \left\| s_\theta(\mathbf{x}_t, t) - \nabla_{x_t} \log q_{0t}(\mathbf x_t | \mathbf x_0) \right\|^2] \label{eq:loss_2}
\end{align}

GraphGDP ~\cite{huang2022graphgdp}, diffuses the graph edges via stochastic differential equations (SDE). For the sampling process tailored to graph data, the authors designed a position-enhanced graph score network employing the structure and position information of graphs and the noisy adjacency matrices. GDSS~\cite{jo2022score} formulated a system of two stochastic differential equations, one for each graph features generation and adjacency matrix generation in parallel for diffusing both nodes and edges. Another SDE based modelGSDM~\cite{luo2022fast}, addressing the problem of Gaussian noise transforming the adjacency matrix into a dense one quickly, restricts the insertion of Gaussian noise to the eigenvalue matrix of the adjacency matrix.

\section{Graph Diffusion models}
\label{gdm}

Multiple works~\cite{kipf2016variational,simonovsky2018graphvae,liu2018constrained,bojchevski2018netgan,de2018molgan,li2018learning,you2018graphrnn,liao2019efficient,dai2020scalable,chen2021order} have demonstrated various graph generation qualities depending on the sequence of node generation. Autoregressive models, in particular, perform optimally~\cite{huang2022graphgdp}. However, they are ineffective in capturing the permutation invariant properties. Diffusion models have come to rescue here, the first score-based model for permutation invariant graph generation, EDP-GNN~\cite{niu2020permutation},  adapts GIN~\cite{xu2018powerful} and employs annealed Langevin dynamic sampling. It assumes undirected graphs, adding Gaussian noise only to the upper triangular part of the adjacency matrix for edge diffusion, and quantizes the continuous adjacency matrix generated when sampling. On predicting edgewise features, EDP-GNNs are empirically shown to be more expressive compared with vanilla ones. Additionally, sample quality is found to be comparable with state-of-the-art models.

GraphGDP~\cite{huang2022graphgdp} also attempted to address the generation of permutation invariant graphs, where state-of-the-art autoregressive models failed. However, the proposed continuous-time generative diffusion model employed stochastic differential equations (SDE) for edge diffusion. Moreover, for sampling, a score network is specifically designed utilizing the structure and position information.

Although diffusion models show great promise, the unique intrinsic properties of graph data call for a few adaptations in the standard diffusion process to make them fit for graph generation tasks. This section next details the modifications thus called for.

\subsection{Discrete Diffusion} 

It is observed that the standard diffusion models have difficulties in capturing the structural properties of the graph data because it relies on a continuous Gaussian noise process which creates fully connected, noisy graphs where structural information is not defined~\cite{ingraham2019generative}. DiGress~\cite{vignac2022digress}, the first discrete graph generation model, features a Markov process noise model where, similar to the way noise is infused independently in each pixel in image diffusion, all progressive noise addition steps are performed independently on each node or edge. The denoising process involves training a graph transformer network to predict the clean graph from a noisy input. It delivers state-of-the-art performance on both molecular and non-molecular datasets. 
Meanwhile, another concurrent work~\cite{haefeli2022diffusion} validates the improvement in the sampling quality and denoising efficiency with the use of discrete noise.

Furthermore, a work~\cite{anand2022protein} in the protein design domain modifies the diffusion process by modeling it as random masking of a portion of the residues. This fraction is decided by linear interpolation between 0 and 1 depending on the time step t. For sampling, the same procedure is reversed, starting with the completely masked residues at time T. These works use the Markov transition probability matrix to implement discrete noise, and the objective becomes a simple cross-entropy loss.

\subsection{Low Rank Diffusion}
For a typical image diffusion model, data is corrupted by adding full-rank Gaussian noise~\cite{ingraham2019generative,dhariwal2021diffusion}. However, diffusing graphs with full-rank noise is detrimental to the training of the score network. In the image diffusion model tasks, the early- and mid-diffusion process stages are not influenced by the inclusion of full-rank Gaussian noise in the image recognition~\cite{song2020improved,song2020score,dhariwal2021diffusion}. Nevertheless, when perturbing the graph with full-rank Gaussian noise, the fatal noise is injected into the unsupported regions and is detrimental to the training of the score network~\cite{alon2020bottleneck,song2020improved,alon2020bottleneck}. It compromises graphic structure and feature representation. To resolve this issue, rather than doing a full-rank diffusion of the adjacency matrix, GSDM~\cite{luo2022fast} restricts the insertion of Gaussian noise to the eigenvalue matrix, which is the spectral decomposition of the adjacency matrix. This approach is effective in both generic graph and molecule generation tasks.

\subsection{Roto-Translation Equivariance and Invariance}

With regards to molecular graph generation specifically, the distribution of molecules should be invariant to rotation and translation, that is, no matter what Euclidean transition is applied to the molecules, their probability distribution should not change~\cite{duvenaud2015convolutional,jorgensen2018deep}. Thus generative models need to make this property true when generating 3D molecular data. The most commonly used method to satisfy these invariances is to employ a roto-translation equivariant score network. ~\cite{xu2022geodiff} proved that an invariant density can be produced by dynamics along an equivariant Gaussian Markov kernel beginning from an invariant standard density. This means that we can ensure the invariance property by making our neural networks equivariant. Consequently, in GeoDiff~\cite{xu2022geodiff}, an equivariant graph convolution network termed as graph field network (GNF) is designed. Inspired by the aforementioned principle, several works~\cite{lin2022diffbp, igashov2022equivariant, hoogeboom2022equivariant, trippe2022diffusion} utilize either equivariant graph neural network~\cite{satorras2021n} or its modification as their score network.  Another trick, is that, zero center of mass (CoM) is also explored for this purpose, as it  moves the center of atomic coordinates to zero, ensuring translation invariance. Another approach, ConfGF~\cite{shi2021learning}, formulates the problem as diffusing interatomic distances instead of atomic coordinates. Based on the observation that those two gradient fields are connected via chain rule, their score networks are roto-translation invariant.

\section{Applications}
\label{app}

\subsection{Molecules}
Deep learning approaches are currently having a critical impact on molecular science, with an increasing number of studies developing deep learning models to analyze or generate molecules~\cite{walters2020applications,atz2021geometric}. These methods can capture high-level structural features and patterns in large data sets by utilizing artificial neural networks, which allows them to generalize across a wide range of molecular structures and properties~\cite{walters2020applications}. For instance, in drug discovery, deep learning models illustrate considerable promise in identifying novel drug candidates with improved effectiveness and decreased side effects~\cite{zhavoronkov2018artificial}.

\subsubsection{Molecular Conformation Generation}

In the field of biological and chemical information, predicting molecular conformations is a fundamental problem~\cite{shi2021learning}. Previous graph generation methods have suffered from several drawbacks that are detrimental to molecular generation. Among them, Markov chain Monte Carlo (MCMC) has high computational cost~\cite{brooks1998markov}; the deep generation model based on intermediate variables leads to implausible generation results due to the errors generated by the multi-stage generation process; and the gradient-based deep generation model violates mathematical theorems.

According to the conventional molecular dynamics simulation force field method, \cite{shi2021learning} presents ConfGF, a unique method for small molecule conformation generation. ConfGF addresses the issue of multistage cumulative error. The gradients of interatomic distances are first estimated by denoising score matching, and then the gradients of coordinates are obtained. They estimate the gradient field of log density of atomic coordinates and directly generate conformation through Langevin dynamics.

Past approaches~\cite{xu2021learning,simm2019generative} generally neglect the interactions between long-range non-bonded atoms. Following the principle of ConfGF~\cite{shi2021learning}, DGSM~\cite{luo2021predicting} uses a score-matching method to directly evaluate the gradient field of the log density of atomic coordinates. They consider not only local interaction between bonded atoms but also long-range interaction between non-bonded atoms. To model long-range interactions in molecules, they augment graph structure dynamically during training and inference. When reconstructing graph structure, they assume that each atom only interacts with all atoms within the cutoff distance, i.e., only atoms within the cut-off distance were connected. The overall structure allows for effective training in an end-to-end manner.

GeoDiff\cite{xu2022geodiff} is a probabilistic model of the generated molecular conformations that considers each atom as a particle. To overcome multi-step errors caused by distances and conformations, GeoDiff~\cite{xu2022geodiff} directly changes the atoms to ensure rotation-translation invariance. ConfGF~\cite{shi2021learning} has accumulated errors in capturing roto-translation invariance using intermediate geometric variables (atom distances) when generating atomic coordinates. Therefore, GeoDiff\cite{xu2022geodiff} restores realistic conformation directly from the noise distribution. It learns the inversion directly by treating the diffusion process as a Markov chain.


The challenge of utilizing diffusion models on molecules lies in that the model does not work by adding the same Gaussian noise to every feature. Consequently, EDM~\cite{hoogeboom2022equivariant}, a novel method for generating 3D molecular structures, uses DDPM to add different noise for various features. EDM employs diffusion processes to sample molecular conformations by transforming initial points in space on both continuous coordinates and categorical atom types.

\begin{figure*}[!htbp]\centering
    \includegraphics[width=0.90\linewidth]{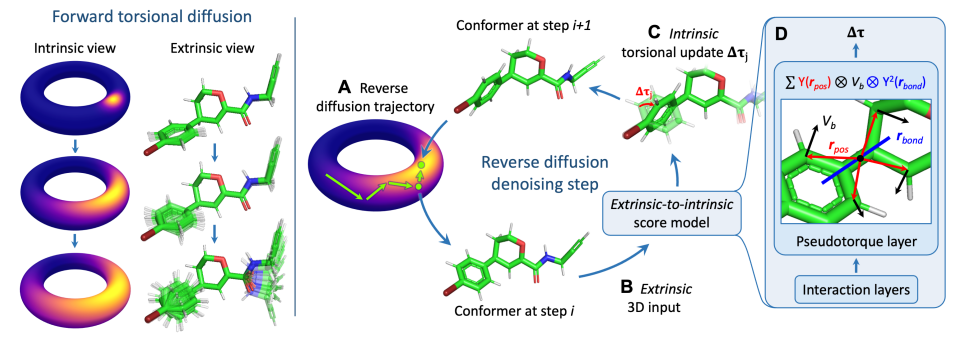}\\
    \caption{Overview of torsional diffusion (figure obtained from~\cite{jing2022torsional})
}
\label{figure_3}
\end{figure*}

Unlike the prior diffusion process on Euclidean space, Torsional diffusion~\cite{jing2022torsional} formulates the diffusion process on the torus defined by torsion angles, as shown in Figure~\ref{figure_3}. The diffusion process over conformers operates only on the torsion angle, with other degrees of freedom fixed, such as bond lengths, bond angles, and cycle conformation. In reverse SDE, they leverage the extrinsic-to-intrinsic score model, which takes the conformer’s 3D structure in Euclidean space as input, and predicts a score on a torsional space to generate a new conformer through an intrinsic torsional update. This has the advantage of being able to generate conformers through two orders of magnitude fewer denoising steps compared to Geodiff~\cite{xu2022geodiff}, sota Euclidean diffusion approach

\subsubsection{Drug Discovery}
Generative diffusion models allow the analysis of large databases of chemical structures and properties and the generation of new compounds optimally targeted for specific drug targets~\cite{jin2018junction,song2020improved}. By harnessing the power of diffusion-based generative models, a broad field of chemistry can be rapidly explored, and promising drug candidates can be identified that would be difficult to discover using traditional methods~\cite{elton2019deep,sanchez2018inverse}. 

A prevalent issue with existing models of molecular generation is the mutual correlation of interatomic distances and interatomic forces. For instance, the interatomic forces are dominated by the van der Waals force when the interatomic distance is less than a certain threshold and chemical bonding dominates the interatomic forces when the distance is greater than a certain threshold. MDM~\cite{huang2022mdm} addresses the issues of capturing the potential of interatomic forces and lack of diversity. MDM encodes the interatomic forces of different strengths by enhancing the potential interatomic forces with a dual equivalence encoder. To augment generative variety, they introduce latent variables in the process of diffusion and generation. MDM outperforms the state-of-the-art model EDM~\cite{hoogeboom2022equivariant}, which is unable to establish chemical bonds through atom pair distances, especially on drug-like datasets.

An effective paradigm for drug design is molecular fragment-based drug discovery, which is the design of linkers between unlinked fragments. DiffLinker~\cite{igashov2022equivariant} is the utilization of diffusion model to design molecular linkers. Conventional molecular linkers connect only pairs of fragments, while DiffLinker allows to ligate any number of fragments. The diffusion framework of DiffLinker is comparable to the EDM~\cite{hoogeboom2022equivariant}, and the denoising model is the same as the EGNN~\cite{wu2022diffusion}. In addition, DiffLinker has an extra module to predict the linker size and automatically determine the atomic number of linkers and their connection points.

Structure-based drug design (SBDD) is the process of designing and optimizing chemical structures with the aim of candidate drug synthesis~\cite{anderson2003process}. DiffSBDD~\cite{schneuing2022structure} generates small molecule ligands with high affinity and particularity according to the protein pocket as a condition. The approach utilizes an equivariant DDPM to generate molecules and binding conformations specific to the protein pocket. The protein and ligand point clouds are represented as fully connected graphs and further processed by EGNNs. The prior and transition probabilities are modeled using an isotropic Gaussian to ensure translation equivariance. DiffSBDD is a powerful tool for generating high-quality ligands for specific protein targets.

It is insufficient to generate accurate molecules with the desired properties by simply using a single conditional generative model, such as EDM~\cite{hoogeboom2022equivariant}. To enable more flexible and controllable generation of molecules, EEGSDE~\cite{bao2022equivariant} formalizes the framework as an equivariant SDE to follow the guidance of the energy function which is carefully designed that the generated molecules are targeted to quantum properties, molecular structures, and even a combination of them in each step. Additionally, gradient descent is performed on energy function so that the generated molecules are encouraged to have low energy. Under the guidance of energy function, EEGSDE~\cite{bao2022equivariant} benefit applications such as drug or material discovery over EDM~\cite{hoogeboom2022equivariant}.(it can be written for section4 because it can be compared by EDM(DDPM-based method).

GCDM~\cite{morehead2023geometry} proposes a novel method for generating 3D molecular structures using a diffusion process that considers the molecule's geometry. The model operates a graph convolutional neural network to encode the 2D structure of a molecule and then uses a diffusion process to generate a 3D structure compatible with the input 2D structure. The method is shown to create high-quality 3D structures that are competitive with state-of-the-art methods while requiring fewer computational resources. The proposed approach has potential applications in drug discovery and material design.

\subsection{Protein Design}

Technically, proteins are also a specical kind of molecules. In cotrast to the proteins discussed above, proteins often have a complex structure which consists of amino acid chains and they play pivotal roles in cellular processes. Many aspects of the protein generation problem have seen good progress but there remain certain challenges~\cite{ingraham2019generative}.

\subsubsection{Structure and Sequence}
The function of a protein is mainly determined by its three-dimensional structure. Considering the importance, one of the key scientific challenges in biotechnology is generating new proteins with specified functional structures and sequences~\cite{alquraishi2019end,senior2020improved,jumper2021highly,yang2020improved}.
Traditional solutions call for expert participation, while, even though machine learning methods relinquished the need for expert help, they lack in generating large structures along with being computationally expensive and necessitating experimental validation~\cite{jumper2021highly}. Moreover, learning spatial ordering of the atoms is requisite to learn a viable chain structure in protein generation tasks, recent diffusion models though successfully generate small molecules do not impose this and consequently are not up to the task.

A number of priors on the distribution of the atom positions constrain the 3D structure of a protein~\cite{anand2022protein} proposes a diffusion based generative model able to construct feasible and diverse proteins. Specifically, to generate feature embeddings given the constraints, a transformer model with triangular self-attention is employed~\cite{mohamed2019transformers}. Taking into consideration the different natures of image and protein data, the attention process is replaced by Invariant Point Attention~\cite{jumper2021highly}. The output of the constraint network and the structure at time t are utilized as the conditions in the modified diffusion process. Specifically, diffusion is done via random masking of residues' fractions, and for sampling the process is reversed. Experiments are conducted for multiple tasks such as 3D design given the protein specification alone, modifying given structures, and joint modeling of structure and sequence showing promising performance~\cite{jumper2021highly,rives2021biological}.

In contrast to modeling the protein backbone via 3D coordinates, a denoising diffusion model~\cite{wu2022protein} mimicking the natural folding process, defines it via six inter-residue angles reflecting the arrangement of amino acid residues. Given the position of the current residue, each vector of these angles defines the relative position of backbone atoms in the subsequent residue. The shift and rotational invariance inherent in this representation eliminates the need for complicated equivariant networks enabling the authors to employ a simple bidirectional transformer architecture. A regression head is used for outputting the angles from the representations learned by the transformer network. Despite showing reasonable performance, the use of angles to represent protein backbone would potentially mean that early errors significantly affect the overall generated structure. Moreover, generated structures are also observed to have collisions.

\subsubsection{Motif conditioned structure design}
Apart from general protein structure design, the design of a stable scaffold/backbone to support a desired structural protein fragment, known as a motif is necessary, particularly, for the design of enzymes, and vaccines
~\cite{trippe2022diffusion}, features ProtDiff, which adapts the equivariant graph neural network~\cite{satorras2021n} for the diffusion of protein backbones. It is the first machine-learning technique, which builds reasonably long protein backbones conditioned on motifs. The conditional sampling of scaffolds is approached as a Sequential Monte Carlo problem in this work, it implements SMCDiff as a particle filtering technique used to sample scaffolds that are more consistent with the motif. The approach is reported to generate valid scaffolds for the target motif. Additionally, unconditional sampling indicated the model’s reasonable capacity to capture the underlying distribution.
\begin{figure*}[!htbp]\centering
    \includegraphics[width=0.90\linewidth]{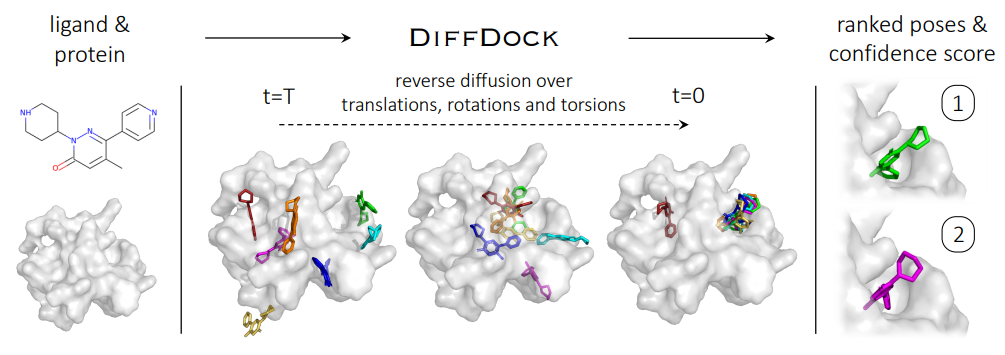}\\
    \caption{Overview of DiffDock (figure obtained from~\cite{corso2022diffdock})
}
\label{figure_4}
\end{figure*}

\subsubsection{Antibody Design}
Another important task in protein generation domain is the design of antibodies, the proteins playing a vital role in the human immune system~\cite{zhou2018deep}. The specificity of antibodies to the antigens is determined by the Complementarity Determining Regions (CDRs)~\cite{north2011new,lefranc2003imgt}. In order to create medicinal antibodies capable of effectively targeting specific antigens, it is crucial to design these CDRs~\cite{dunbar2014sabdab}. 
Traditional methods involved complex energy functions and thus had been inefficient and prone to providing suboptimal solutions. Moreover, though recent deep learning methods have shown improved performance, these are still limited in their capacity to generate antibodies based on previously observed ones only. It is identified in~\cite{luo2022antigen} that a number of challenges for practical use scenario, that is, a generative model should output  CDRs compatible with the 3D structure of antigen, consider position and orientation of amino acids, and given an antibody, be able to optimize its binding affinity. To this end,~\cite{luo2022antigen} proposes the first deep learning model DiffAb to perform joint design of antibody sequence and structure conditioned on the 3D structures of the antigen. It takes an antigen-antibody complex as input and diffuses the amino acid type, position, and orientation on CDRs.

\subsubsection{Molecular Docking}
Novel biological discoveries can benefit hugely from the advances in the generation of protein binding molecules, called ligands, and 3D structure of protein-ligand molecular complex~\cite{mobley2009binding}. DiffBP~\cite{lin2022diffbp}, a DDPM model, is proposed for generating ligands, and to account for the fact that in the molecular systems, the interactions among atoms are influenced by all other atoms. They generate molecules at the full atom level. For full-atom generation, though, it is necessary to assign the atom numbers of the binding molecules before diffusion and denoising process. Therefore, in the pre-generation process before the denoising, they use additional GNN that generates the atom numbers and the molecule’s CoM, and after that, denoise both 3D coordinates and element type of a molecule with an SE(3)-EGNN. DiffBP~\cite{lin2022diffbp} can generate binding molecules with proper size, higher affinity with target protein, and roto-translation invariances. Unlike these methods, DiffDock~\cite{corso2022diffdock} defines diffusion process over ligand poses(assignment of atomic positions in R3) corresponding to the degrees of freedom involved in molecular docking: ligand translations, ligand rotations, torsion angles. Additionally, they have high selective accuracy by leveraging the confidence model that predicts confidence estimates on ligand poses generated through reverse SDE. The overview of DiffDock is shown in Figure~\ref{figure_4}. Moreover, for the protein-ligand complex generation,  NeuralPLexer~\cite{qiao2022dynamic}, an SDE-based generative model conditioned on the protein backbone templates and ligand molecular graphs is proposed. Particularly, the encoded inputs are forwarded to a contact predictor, which iteratively samples the distributions for binding interface proximity for each ligand. An SDE parameterized by the previously produced contact maps adds noise to the data.  Protein and ligand structures are then jointly generated with an equivariant structure diffusion module (ESDM) by performing reverse diffusion of atom coordinates.

\subsection{Others}

\begin{figure*}[!htbp]\centering
    \includegraphics[width=0.80\linewidth]{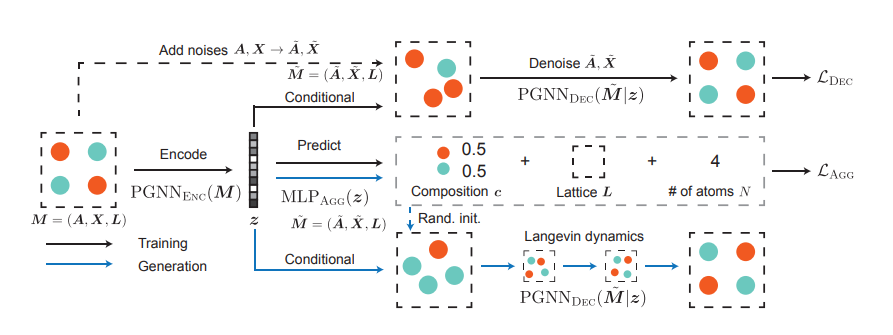}\\
    \caption{Overview of CDVAE approach (figure obtained from~\cite{xie2021crystal})
}
\label{figure_5}
\end{figure*}

\textbf{Materials Design.}
Material is represented as the periodic arrangement of atoms in the 3D space~\cite{glusker1996crystal}. Specifically, the Unit cell represents the infinite periodic structure of a material, which in turn is characterized by atom types, coordinates, and the periodic lattice. To generate stable materials, there are two key points: requirements 1) atom coordinates of material must exist in the local energy minimum defined by quantum mechanics, and requirements 2)The structure follows specific bonding preferences between different atom types. Moreover, the structure of the material should be permutation, translation, rotation, and periodic invariant. CDVAE~\cite{xie2021crystal} first encodes material onto a lower dimensional latent space, and then MLP property predictor predicts the number of atoms, periodic lattice, and the composition from latent representation z. Periodic GNN decoder, noise conditional score network(NCSN)~\cite{song2019generative}, denoise atom type, and atom coordinate from z. The decoder outputs gradients leading atom coordinates to the energy local minimum and updates atom types to capture the specific local boding preferences based on the neighbors. CDVAE~\cite{xie2021crystal} learns stability in making stable materials unstable by adding noise and then gradually denoising them into a stable material. The overview of CDVAE is shown in Figure~\ref{figure_5}.

\textbf{Mischaneous.}
Early work in graph diffusion diffuses the adjacency matrix for edge diffusion, EDP-GNN~\cite{niu2020permutation}, focusing on undirected graph generation adds noise only to the upper triangular part of adjacency matrix. Then, for making generated graphs discrete at sampling time, they threshold the sampled continuous adjacency matrix. Although methods diffusing only one structural element of graphs show promising results, they do not completely capture the relationship between nodes and edges, which is important for generating real-world graphs such as molecules. To overcome this issue, GDSS~\cite{jo2022score} diffuses both via two SDEs. It models the complex dependencies between nodes and edges in parallel. Since nodes and edges behave differently in terms of permutation, different modeling is desired, thus GDSS uses graph attention network for the denoising process. However, models learning node and edge representations independently ignore the interdependency between them~\cite{chen2022nvdiff}. To address these issues, NVDiff~\cite{chen2022nvdiff} diffuses only the latent representation of node vectors encoded via GraphVAE~\cite{kipf2016variational}, capturing the structural information of graph, and uses an attention-based score generative model (SGM) as a prior to capture the local and global context within the graph.

\begin{table*}[!htbp] \centering 
\caption{Results on the Community-small and Ego-small datasets}
\label{tab:simplegraphs}
\resizebox{0.8\textwidth}{!}{
\begin{tabular}{c|c|c|c|c|c|c|c|c}
\hline
\textbf{Method} & \multicolumn{8}{c}{\textbf{Datasets}} \\
\cline{2-9}
 & \multicolumn{4}{c|}{\textbf{Community-small}} & \multicolumn{4}{c}{\textbf{Ego-small}} \\
\cline{2-9}
& \textbf{Deg.} & \textbf{Clus.} & \textbf{Orbit} & \textbf{Avg.}
 & \textbf{Deg.} & \textbf{Clus.} & \textbf{Orbit} & \textbf{Avg.} \\

\hline
EDP-GNN~\cite{niu2020permutation} &  0.053 & 0.144  & 0.026  &  0.074  &  0.052 & 0.093  & 0.007  & 0.050  \\
\hline
GDSS~\cite{jo2022score} & 0.021  & 0.024   & 0.007  & 0.017  &  0.045 & 0.086  & 0.007  &  0.046 \\
\hline
GSDM~\cite{luo2022fast} &   0.011 & 0.015  & 0.001  & 0.009  & -  & -  & -  & -  \\
\hline
NVDiff~\cite{chen2022nvdiff} & 0.021  & 0.035  & 0.018  & 0.025  & 0.005  & 0.045 & 0.001  &  0.017 \\
\hline
GraphGDP~\cite{huang2022graphgdp} & 0.039  & 0.074  & -  &  - &  0.023 & 0.029  & -  & -  \\
\hline
\end{tabular}}
\end{table*}

\section{Evaluation} \label{eva}
Validating the effectiveness of generative models entails that both generated data and the learned distributions be evaluated~\cite{salimans2016improved,sutherland2016generative,theis2015note}. For graph data, this is twofold difficult. First, unlike prediction problems which only need an evaluation of deterministic predictions, generative tasks necessitate evaluating the learned distributions~\cite{sajjadi2018assessing,theis2015note}. Second, evaluating graph-structured data is more complicated than assessing image or text data~\cite{zhou2020graph,hamilton2017inductive}. We summarize the most generally operated graph generation metrics, including domain-independent and application-specific graph generation~\cite{zhang2019graph}.

\begin{table*}[!htbp] \centering 
\caption{Results on the QM9 and ZINC250K datasets}
\label{tab:complicatedgraphs}
\resizebox{0.8\textwidth}{!}{
\begin{tabular}{c|c|c|c|c|c|c}
\hline
\textbf{Method} & \multicolumn{6}{c}{\textbf{Datasets}} \\
\cline{2-7}
 & \multicolumn{3}{c|}{\textbf{QM9}} & \multicolumn{3}{c}{\textbf{ZINC250K}} \\
\cline{2-7}
& \textbf{Val↑} & \textbf{NSPDK↓} & \textbf{FCD↓} & \textbf{Val↑}& \textbf{NSPDK↓} & \textbf{FCD↓}  \\

\hline
EDP-GNN~\cite{niu2020permutation} &  47.52 &0.005 & 2.680 &  82.97 &  0.049 & 16.737\\
\hline
GDSS~\cite{jo2022score} & 95.72  & 0.003 & 2.900 & 97.01 &  0.019& 14.656\\
\hline
GSDM~\cite{luo2022fast} &   99.9 & 0.003  & 2.650  & 92.70 & 0.017  & 12.956 \\
\hline
NVDiff~\cite{chen2022nvdiff} & 95.79  & -  & 1.131 & 85.63 & - & 4.019 \\
\hline

\end{tabular}}
\end{table*}

\noindent
\textbf{Datasets.}
Though there are many graph generation datasets, more specific to the target tasks, yet most representative ones include 2D datasets, that is, Community-small, Ego-small, and Grid and 3D datasets, such as QM9, ZINC250k, and Enzymes as shown in Table~\ref{tab:alldatasets}. In detail, Community-small consists of 100 small community graphs with 12 to 20 nodes and Ego-small is a set of 200 citation ego graphs consisting of 4 to 18 nodes. The number of nodes in Grid, a dataset of 100 grid graphs, goes up to 400. Molecule generation datasets, QM9 and ZINC250k, contain more than a hundred thousand molecular graphs with the later comprising graphs as large as 38 nodes. Enzymes dataset, used in protein generation tasks, contains 600 protein graphs with the number of nodes ranging from 10 to 125. 

\begin{table*}[!htbp] \centering 
\caption{Datasets commonly used for graph generation models}
\label{tab:alldatasets}
\resizebox{0.9\textwidth}{!}{
\begin{tabular}{c|c|c|c|c}
\hline
\textbf{Dataset}  &  \textbf{Dimensionality}  &  \textbf{Category}   &  \textbf{No. of Graphs (G)} & \textbf{No. of Nodes (N)} \\
\hline
Community-small & 2D & Social & 100 & 11$<$N$<$20\\
\hline
Ego-small & 2D & Social & 200 & 3$<$N$<$18\\
\hline
Grid & 2D & Grid & 100 & N$<=$400\\
\hline
QM9 & 3D & Bioinformatics/Molecular & 130,831 & 3$<$N$<$29\\
\hline
ZINC250K & 3D & Bioinformatics/Molecular & 249,456 & 6$<$N$<$38\\
\hline  
Enzymes & 3D & Bioinformatics/Protein & 600 & 9$<$N$<$125\\
\hline  
\end{tabular}}
\end{table*} 

\noindent
\textbf{Metrics.}

Most represented types of evaluation metrics in the graph generation literature are statistical, classifier-based, and related to quality. Evidently, \textbf{statistical} evaluations compare the statistics’ distributions of the generated graphs to that of real graphs via distance metrics such as Average Kullback-Leibler Divergence, Maximum Mean Discrepancy, etc. There are seven typical graph statistics~\cite{clauset2004finding,karrer2011stochastic}, including node degree distribution (Deg.) informing about the graph’s local connectivity patterns, clustering coefficient distribution (Clus.) measured as the ratio of a node’s potential number of triangles to the actual orbit count distribution (Orbit), the distribution of the 4-node orbit counts, useful in understanding the model’s capability to match higher-order graph statistics, and more. Whereas, in \textbf{Classifier-based} evaluation, a Graph Isomorphism Network (GIN) classifier trained on real graphs is tested on generated graphs via measuring either the classification accuracy of generated graphs or the Fréchet Inception Distance (FID) between the embeddings of generated and real graphs. However, as much as evaluating the similarity of real and generated graphs is critical, many applications like molecule generation demand some measure of \textbf{quality}. One is validity, which assesses whether certain requirements criteria, such as structural properties like being a tree or cycle, or chemical properties for molecule generation, are met. In Table~\ref{tab:simplegraphs} and Table~\ref{tab:complicatedgraphs}, we summarize the results of some representative diffusion models tested in the graph domain. \textbf{Uniqueness} is another such metric; the fraction of graphs that are not sub-graph isomorphic to other generated graphs are computed to capture the diversity, key to some tasks, for instance, drug design. Novelty, on the other hand, measures the same but with regards to the real graphs to indicate generalizability.

\section{Challenges} \label{ch}
Graph generation via diffusion models is a relatively emerging research area. Though several works have employed diffusion models for graph generation, the field still faces many challenges and open research questions. In this section, we outline some of the foremost challenges.

\textbf{Evaluation Criterion.}
Despite various widely adopted evaluation metrics, measuring and comparing the performance of graph generative models remains a major challenge given the lack of ground truth and universal metrics~\cite{xu2018graph2seq}. Current evaluations employing prior knowledge about graph data, for example, statistics and properties can not be relied on in real-world applications, in particular molecular generation, often requiring more expensive evaluations. The importance of these statistics and properties also varies with the target task. Thus, further research is needed to develop appropriate evaluation metrics.

\textbf{Diversity of Graphs.}
Graphs are prevalent in many applications and a variety of data can be modeled via them. Each type of graph has unique characteristics and properties, which require different modeling approaches to generate them effectively. For instance, social networks have specific features, such as community structures and degree distributions, which must be considered when generating them~\cite{yang2016revisiting,wang2016structural}. Similarly, biological networks require specialized models that incorporate the biochemical and physiological properties of the system being studied~\cite{ingraham2019generative}. Despite this diversity of graph data, the current research on graph generation primarily focuses on molecular graphs used in drug discovery and materials science~\cite{gomez2018automatic}. One of the reasons for this is the availability of many repositories of molecular graphs, such as the PubChem and ChEMBL databases. However, different graphs typically have significantly distinct natures and may need generation models to fulfill specific criteria accordingly. Despite this variety of graph data, research is mainly aimed at molecular graphs, partly due to the lack of ample repositories, overlooking the diversity of real graph data.

\textbf{Scalability.}
Largely practiced diffusion of edges is not feasible for huge real-world graphs, as it necessitates a reverse diffusion of quadratically increasing steps for sampling new graphs. The cost of generating a new graph by diffusion increases dramatically when the number of nodes grows~\cite{ying2018graph}. This computational bottleneck makes it challenging to generate large-scale graphs using diffusion-based methods, limiting their application to large graphs common in the real world. Though efforts have been made to alleviate the problem by diffusing latent representation of nodes instead, the graph generation via diffusion is far from scalable and needs further exploration.

\textbf{Irregularity.}
Graphs can have a highly irregular structure with variable numbers of nodes and edges, which can pose difficulties in specifying a diffusion process that can effectively seize the dynamics of the graph~\cite{velivckovic2017graph,velickovic2019deep}. In particular, designing a diffusion process that captures the complex dependencies between nodes and edges in a graph is challenging~\cite{perozzi2014deepwalk}. Additionally, the irregular structure of graphs can make it difficult to define a fixed diffusion kernel or transition operator suitable for modeling the dynamics of the graph~\cite{zhang2019graph}. As a result, adapting diffusion models for graph data requires developing new techniques for incorporating the irregular structure of graphs into the diffusion process, which is an active area of research.

\textbf{Interpretability.}
Diffusion models depend on a score function to calculate the probability density of the data. However, the score function can be challenging to interpret and visualize in the context of graphs, which can restrict the model's interpretability and make it difficult to achieve insights into the underlying dynamics of the graphs~\cite{you2018graphrnn,ingraham2019generative}. As a result, analyzing and comprehending the behavior of diffusion models in the context of graphs can be challenging.

\section{Conclusion}

This work conducts a review of the development of graph generation in generative AI. Specifically, we focus on its recent progress with graph diffusion models. Our survey wok can help readers have a quick yet comprehensive understanding of graph diffusion models.

\bibliographystyle{IEEEtran}
\bibliography{bib_mixed,bib_local}

\begin{thebibliography}{100}
\providecommand{\url}[1]{#1}
\csname url@samestyle\endcsname
\providecommand{\newblock}{\relax}
\providecommand{\bibinfo}[2]{#2}
\providecommand{\BIBentrySTDinterwordspacing}{\spaceskip=0pt\relax}
\providecommand{\BIBentryALTinterwordstretchfactor}{4}
\providecommand{\BIBentryALTinterwordspacing}{\spaceskip=\fontdimen2\font plus
\BIBentryALTinterwordstretchfactor\fontdimen3\font minus
  \fontdimen4\font\relax}
\providecommand{\BIBforeignlanguage}[2]{{%
\expandafter\ifx\csname l@#1\endcsname\relax
\typeout{** WARNING: IEEEtran.bst: No hyphenation pattern has been}%
\typeout{** loaded for the language `#1'. Using the pattern for}%
\typeout{** the default language instead.}%
\else
\language=\csname l@#1\endcsname
\fi
#2}}
\providecommand{\BIBdecl}{\relax}
\BIBdecl

\bibitem{sun2009concise}
J.~Sun, M.~Ovsjanikov, and L.~Guibas, ``A concise and provably informative
  multi-scale signature based on heat diffusion,'' in \emph{Computer graphics
  forum}, vol.~28, no.~5.\hskip 1em plus 0.5em minus 0.4em\relax Wiley Online
  Library, 2009, pp. 1383--1392.

\bibitem{mikolov2013efficient}
T.~Mikolov, K.~Chen, G.~Corrado, and J.~Dean, ``Efficient estimation of word
  representations in vector space,'' \emph{arXiv preprint arXiv:1301.3781},
  2013.

\bibitem{kipf2016semi}
T.~N. Kipf and M.~Welling, ``Semi-supervised classification with graph
  convolutional networks,'' \emph{arXiv preprint arXiv:1609.02907}, 2016.

\bibitem{hamilton2017inductive}
W.~Hamilton, Z.~Ying, and J.~Leskovec, ``Inductive representation learning on
  large graphs,'' \emph{Advances in neural information processing systems},
  vol.~30, 2017.

\bibitem{chen2019graph}
C.~Chen, W.~Ye, Y.~Zuo, C.~Zheng, and S.~P. Ong, ``Graph networks as a
  universal machine learning framework for molecules and crystals,''
  \emph{Chemistry of Materials}, vol.~31, no.~9, pp. 3564--3572, 2019.

\bibitem{hamaguchi2017knowledge}
T.~Hamaguchi, H.~Oiwa, M.~Shimbo, and Y.~Matsumoto, ``Knowledge transfer for
  out-of-knowledge-base entities: A graph neural network approach,''
  \emph{arXiv preprint arXiv:1706.05674}, 2017.

\bibitem{bacciu2020gentle}
D.~Bacciu, F.~Errica, A.~Micheli, and M.~Podda, ``A gentle introduction to deep
  learning for graphs,'' \emph{Neural Networks}, vol. 129, pp. 203--221, 2020.

\bibitem{duvenaud2015convolutional}
D.~K. Duvenaud, D.~Maclaurin, J.~Iparraguirre, R.~Bombarell, T.~Hirzel,
  A.~Aspuru-Guzik, and R.~P. Adams, ``Convolutional networks on graphs for
  learning molecular fingerprints,'' vol.~28, 2015.

\bibitem{bruna2013spectral}
J.~Bruna, W.~Zaremba, A.~Szlam, and Y.~LeCun, ``Spectral networks and locally
  connected networks on graphs,'' \emph{arXiv preprint arXiv:1312.6203}, 2013.

\bibitem{kyrola2012graphchi}
A.~Kyrola, G.~Blelloch, and C.~Guestrin, ``Graphchi: Large-scale graph
  computation on just a $\{$PC$\}$,'' in \emph{Presented as part of the 10th
  $\{$USENIX$\}$ Symposium on Operating Systems Design and Implementation
  ($\{$OSDI$\}$ 12)}, 2012, pp. 31--46.

\bibitem{malewicz2010pregel}
G.~Malewicz, M.~H. Austern, A.~J. Bik, J.~C. Dehnert, I.~Horn, N.~Leiser, and
  G.~Czajkowski, ``Pregel: a system for large-scale graph processing,'' in
  \emph{Proceedings of the 2010 ACM SIGMOD International Conference on
  Management of data}, 2010, pp. 135--146.

\bibitem{gonzalez2012powergraph}
J.~E. Gonzalez, Y.~Low, H.~Gu, D.~Bickson, and C.~Guestrin, ``Powergraph:
  Distributed graph-parallel computation on natural graphs,'' in
  \emph{Presented as part of the 10th $\{$USENIX$\}$ Symposium on Operating
  Systems Design and Implementation ($\{$OSDI$\}$ 12)}, 2012, pp. 17--30.

\bibitem{hamilton2017representation}
W.~L. Hamilton, R.~Ying, and J.~Leskovec, ``Representation learning on graphs:
  Methods and applications,'' \emph{arXiv preprint arXiv:1709.05584}, 2017.

\bibitem{zhang2020deep}
Z.~Zhang, P.~Cui, and W.~Zhu, ``Deep learning on graphs: A survey,'' \emph{IEEE
  Transactions on Knowledge and Data Engineering}, vol.~34, no.~1, pp.
  249--270, 2020.

\bibitem{wang2019heterogeneous}
X.~Wang, H.~Ji, C.~Shi, B.~Wang, Y.~Ye, P.~Cui, and P.~S. Yu, ``Heterogeneous
  graph attention network,'' in \emph{The world wide web conference}, 2019, pp.
  2022--2032.

\bibitem{chiang2019cluster}
W.-L. Chiang, X.~Liu, S.~Si, Y.~Li, S.~Bengio, and C.-J. Hsieh, ``Cluster-gcn:
  An efficient algorithm for training deep and large graph convolutional
  networks,'' in \emph{Proceedings of the 25th ACM SIGKDD international
  conference on knowledge discovery \& data mining}, 2019, pp. 257--266.

\bibitem{zhu2020beyond}
J.~Zhu, Y.~Yan, L.~Zhao, M.~Heimann, L.~Akoglu, and D.~Koutra, ``Beyond
  homophily in graph neural networks: Current limitations and effective
  designs,'' \emph{Advances in Neural Information Processing Systems}, vol.~33,
  pp. 7793--7804, 2020.

\bibitem{wu2018moleculenet}
Z.~Wu, B.~Ramsundar, E.~N. Feinberg, J.~Gomes, C.~Geniesse, A.~S. Pappu,
  K.~Leswing, and V.~Pande, ``Moleculenet: a benchmark for molecular machine
  learning,'' \emph{Chemical science}, vol.~9, no.~2, pp. 513--530, 2018.

\bibitem{fout2017protein}
A.~Fout, J.~Byrd, B.~Shariat, and A.~Ben-Hur, ``Protein interface prediction
  using graph convolutional networks,'' \emph{Advances in neural information
  processing systems}, vol.~30, 2017.

\bibitem{ingraham2019generative}
J.~Ingraham, V.~Garg, R.~Barzilay, and T.~Jaakkola, ``Generative models for
  graph-based protein design,'' \emph{Advances in neural information processing
  systems}, vol.~32, 2019.

\bibitem{popova2018deep}
M.~Popova, O.~Isayev, and A.~Tropsha, ``Deep reinforcement learning for de novo
  drug design,'' \emph{Science advances}, vol.~4, no.~7, p. eaap7885, 2018.

\bibitem{bjerrum2017molecular}
E.~J. Bjerrum and R.~Threlfall, ``Molecular generation with recurrent neural
  networks (rnns),'' \emph{arXiv preprint arXiv:1705.04612}, 2017.

\bibitem{jing2018deep}
Y.~Jing, Y.~Bian, Z.~Hu, L.~Wang, and X.-Q.~S. Xie, ``Deep learning for drug
  design: an artificial intelligence paradigm for drug discovery in the big
  data era,'' \emph{The AAPS journal}, vol.~20, pp. 1--10, 2018.

\bibitem{li2015gated}
Y.~Li, D.~Tarlow, M.~Brockschmidt, and R.~Zemel, ``Gated graph sequence neural
  networks,'' \emph{arXiv preprint arXiv:1511.05493}, 2015.

\bibitem{you2018graphrnn}
J.~You, R.~Ying, X.~Ren, W.~Hamilton, and J.~Leskovec, ``Graphrnn: Generating
  realistic graphs with deep auto-regressive models,'' in \emph{International
  conference on machine learning}.\hskip 1em plus 0.5em minus 0.4em\relax PMLR,
  2018, pp. 5708--5717.

\bibitem{popova2019molecularrnn}
M.~Popova, M.~Shvets, J.~Oliva, and O.~Isayev, ``Molecularrnn: Generating
  realistic molecular graphs with optimized properties,'' \emph{arXiv preprint
  arXiv:1905.13372}, 2019.

\bibitem{simonovsky2018graphvae}
M.~Simonovsky and N.~Komodakis, ``Graphvae: Towards generation of small graphs
  using variational autoencoders,'' in \emph{International conference on
  artificial neural networks}.\hskip 1em plus 0.5em minus 0.4em\relax Springer,
  2018, pp. 412--422.

\bibitem{de2018molgan}
N.~De~Cao and T.~Kipf, ``Molgan: An implicit generative model for small
  molecular graphs,'' \emph{arXiv preprint arXiv:1805.11973}, 2018.

\bibitem{yang2022diffusion}
R.~Yang, P.~Srivastava, and S.~Mandt, ``Diffusion probabilistic modeling for
  video generation,'' \emph{arXiv preprint arXiv:2203.09481}, 2022.

\bibitem{ulhaq2022efficient}
A.~Ulhaq, N.~Akhtar, and G.~Pogrebna, ``Efficient diffusion models for vision:
  A survey,'' \emph{arXiv preprint arXiv:2210.09292}, 2022.

\bibitem{cao2022survey}
H.~Cao, C.~Tan, Z.~Gao, G.~Chen, P.-A. Heng, and S.~Z. Li, ``A survey on
  generative diffusion model,'' \emph{arXiv preprint arXiv:2209.02646}, 2022.

\bibitem{zhang2023text}
C.~Zhang, C.~Zhang, M.~Zhang, and I.~S. Kweon, ``Text-to-image diffusion models
  in generative ai: A survey,'' \emph{arXiv preprint arXiv:2303.07909}, 2023.

\bibitem{zhang2023audio}
C.~Zhang, C.~Zhang, S.~Zheng, M.~Zhang, M.~Qamar, S.-H. Bae, and I.~S. Kweon,
  ``A survey on audio diffusion models: Text to speech synthesis and
  enhancement in generative ai,'' \emph{arXiv preprint arXiv:2303.13336}, 2023.

\bibitem{zhang2023complete}
C.~Zhang, C.~Zhang, S.~Zheng, Y.~Qiao, C.~Li, M.~Zhang, S.~K. Dam, C.~M. Thwal,
  Y.~L. Tun, L.~L. Huy, D.~kim, S.-H. Bae, L.-H. Lee, Y.~Yang, H.~T. Shen,
  I.~S. Kweon, and C.~S. Hong, ``A complete survey on generative ai (aigc): Is
  chatgpt from gpt-4 to gpt-5 all you need?'' \emph{arXiv preprint
  arXiv:2303.11717}, 2023.

\bibitem{zhang2023ChatGPT}
C.~Zhang, C.~Zhang, C.~Li, S.~Zheng, Y.~Qiao, S.~K. Dam, M.~Zhang, J.~U. Kim,
  S.~T. Kim, G.-M. Park, J.~Choi, S.-H. Bae, L.-H. Lee, P.~Hui, I.~S. Kweon,
  and C.~S. Hong, ``One small step for generative ai, one giant leap for agi: A
  complete survey on chatgpt in aigc era,'' \emph{researchgate
  DOI:10.13140/RG.2.2.24789.70883}, 2023.

\bibitem{zhang2022survey}
C.~Zhang, C.~Zhang, J.~Song, J.~S.~K. Yi, K.~Zhang, and I.~S. Kweon, ``A survey
  on masked autoencoder for self-supervised learning in vision and beyond,''
  \emph{arXiv preprint arXiv:2208.00173}, 2022.

\bibitem{erdHos1960evolution}
P.~Erd{\H{o}}s, A.~R{\'e}nyi \emph{et~al.}, ``On the evolution of random
  graphs,'' \emph{Publ. Math. Inst. Hung. Acad. Sci}, vol.~5, no.~1, pp.
  17--60, 1960.

\bibitem{watts1998collective}
D.~J. Watts and S.~H. Strogatz, ``Collective dynamics of
  ‘small-world’networks,'' \emph{nature}, vol. 393, no. 6684, pp. 440--442,
  1998.

\bibitem{barabasi1999emergence}
A.-L. Barab{\'a}si and R.~Albert, ``Emergence of scaling in random networks,''
  \emph{science}, vol. 286, no. 5439, pp. 509--512, 1999.

\bibitem{chakrabarti2004r}
D.~Chakrabarti, Y.~Zhan, and C.~Faloutsos, ``R-mat: A recursive model for graph
  mining,'' in \emph{Proceedings of the 2004 SIAM International Conference on
  Data Mining}.\hskip 1em plus 0.5em minus 0.4em\relax SIAM, 2004, pp.
  442--446.

\bibitem{leskovec2010kronecker}
J.~Leskovec, D.~Chakrabarti, J.~Kleinberg, C.~Faloutsos, and Z.~Ghahramani,
  ``Kronecker graphs: an approach to modeling networks.'' \emph{Journal of
  Machine Learning Research}, vol.~11, no.~2, 2010.

\bibitem{bojchevski2018netgan}
A.~Bojchevski, O.~Shchur, D.~Z{\"u}gner, and S.~G{\"u}nnemann, ``Netgan:
  Generating graphs via random walks,'' in \emph{International conference on
  machine learning}.\hskip 1em plus 0.5em minus 0.4em\relax PMLR, 2018, pp.
  610--619.

\bibitem{yang2016revisiting}
Z.~Yang, W.~Cohen, and R.~Salakhudinov, ``Revisiting semi-supervised learning
  with graph embeddings,'' in \emph{International conference on machine
  learning}.\hskip 1em plus 0.5em minus 0.4em\relax PMLR, 2016, pp. 40--48.

\bibitem{vaswani2017attention}
A.~Vaswani, N.~Shazeer, N.~Parmar, J.~Uszkoreit, L.~Jones, A.~N. Gomez,
  L.~Kaiser, and I.~Polosukhin, ``Attention is all you need,'' in
  \emph{NeurIPS}, 2017.

\bibitem{kingma2013auto}
D.~P. Kingma and M.~Welling, ``Auto-encoding variational bayes,'' \emph{arXiv
  preprint arXiv:1312.6114}, 2013.

\bibitem{dinh2016density}
L.~Dinh, J.~Sohl-Dickstein, and S.~Bengio, ``Density estimation using real
  nvp,'' \emph{arXiv preprint arXiv:1605.08803}, 2016.

\bibitem{goodfellow2020generative}
I.~Goodfellow, J.~Pouget-Abadie, M.~Mirza, B.~Xu, D.~Warde-Farley, S.~Ozair,
  A.~Courville, and Y.~Bengio, ``Generative adversarial networks,''
  \emph{Communications of the ACM}, vol.~63, no.~11, pp. 139--144, 2020.

\bibitem{sutskever2014sequence}
I.~Sutskever, O.~Vinyals, and Q.~V. Le, ``Sequence to sequence learning with
  neural networks,'' \emph{Advances in neural information processing systems},
  vol.~27, 2014.

\bibitem{radford2017learning}
A.~Radford, R.~Jozefowicz, and I.~Sutskever, ``Learning to generate reviews and
  discovering sentiment,'' \emph{arXiv preprint arXiv:1704.01444}, 2017.

\bibitem{reed2017parallel}
S.~Reed, A.~Oord, N.~Kalchbrenner, S.~G. Colmenarejo, Z.~Wang, Y.~Chen,
  D.~Belov, and N.~Freitas, ``Parallel multiscale autoregressive density
  estimation,'' in \emph{International Conference on Machine Learning}.\hskip
  1em plus 0.5em minus 0.4em\relax PMLR, 2017, pp. 2912--2921.

\bibitem{gilmer2017neural}
J.~Gilmer, S.~S. Schoenholz, P.~F. Riley, O.~Vinyals, and G.~E. Dahl, ``Neural
  message passing for quantum chemistry,'' 2017, pp. 1263--1272.

\bibitem{ying2018graph}
R.~Ying, R.~He, K.~Chen, P.~Eksombatchai, W.~L. Hamilton, and J.~Leskovec,
  ``Graph convolutional neural networks for web-scale recommender systems,'' in
  \emph{Proceedings of the 24th ACM SIGKDD international conference on
  knowledge discovery \& data mining}, 2018, pp. 974--983.

\bibitem{shi2020graphaf}
C.~Shi, M.~Xu, Z.~Zhu, W.~Zhang, M.~Zhang, and J.~Tang, ``Graphaf: a flow-based
  autoregressive model for molecular graph generation,'' \emph{arXiv preprint
  arXiv:2001.09382}, 2020.

\bibitem{bui2018neural}
T.~D. Bui, S.~Ravi, and V.~Ramavajjala, ``Neural graph learning: Training
  neural networks using graphs,'' in \emph{Proceedings of the Eleventh ACM
  International Conference on Web Search and Data Mining}, 2018, pp. 64--71.

\bibitem{you2018graph}
J.~You, B.~Liu, Z.~Ying, V.~Pande, and J.~Leskovec, ``Graph convolutional
  policy network for goal-directed molecular graph generation,'' \emph{Advances
  in neural information processing systems}, vol.~31, 2018.

\bibitem{doersch2016tutorial}
C.~Doersch, ``Tutorial on variational autoencoders,'' \emph{arXiv preprint
  arXiv:1606.05908}, 2016.

\bibitem{ma2018constrained}
T.~Ma, J.~Chen, and C.~Xiao, ``Constrained generation of semantically valid
  graphs via regularizing variational autoencoders,'' \emph{Advances in Neural
  Information Processing Systems}, vol.~31, 2018.

\bibitem{grover2019graphite}
A.~Grover, A.~Zweig, and S.~Ermon, ``Graphite: Iterative generative modeling of
  graphs,'' in \emph{International conference on machine learning}.\hskip 1em
  plus 0.5em minus 0.4em\relax PMLR, 2019, pp. 2434--2444.

\bibitem{liu2018constrained}
Q.~Liu, M.~Allamanis, M.~Brockschmidt, and A.~Gaunt, ``Constrained graph
  variational autoencoders for molecule design,'' \emph{Advances in neural
  information processing systems}, vol.~31, 2018.

\bibitem{jin2018junction}
W.~Jin, R.~Barzilay, and T.~Jaakkola, ``Junction tree variational autoencoder
  for molecular graph generation,'' 2018, pp. 2323--2332.

\bibitem{kipf2016variational}
T.~N. Kipf and M.~Welling, ``Variational graph auto-encoders,'' \emph{arXiv
  preprint arXiv:1611.07308}, 2016.

\bibitem{rezende2015variational}
D.~Rezende and S.~Mohamed, ``Variational inference with normalizing flows,'' in
  \emph{International conference on machine learning}.\hskip 1em plus 0.5em
  minus 0.4em\relax PMLR, 2015, pp. 1530--1538.

\bibitem{papamakarios2017masked}
G.~Papamakarios, T.~Pavlakou, and I.~Murray, ``Masked autoregressive flow for
  density estimation,'' \emph{Advances in neural information processing
  systems}, vol.~30, 2017.

\bibitem{ho2019flow++}
J.~Ho, X.~Chen, A.~Srinivas, Y.~Duan, and P.~Abbeel, ``Flow++: Improving
  flow-based generative models with variational dequantization and architecture
  design,'' in \emph{International Conference on Machine Learning}.\hskip 1em
  plus 0.5em minus 0.4em\relax PMLR, 2019, pp. 2722--2730.

\bibitem{madhawa2019graphnvp}
K.~Madhawa, K.~Ishiguro, K.~Nakago, and M.~Abe, ``Graphnvp: An invertible flow
  model for generating molecular graphs,'' \emph{arXiv preprint
  arXiv:1905.11600}, 2019.

\bibitem{liu2019graph}
J.~Liu, A.~Kumar, J.~Ba, J.~Kiros, and K.~Swersky, ``Graph normalizing flows,''
  \emph{Advances in Neural Information Processing Systems}, vol.~32, 2019.

\bibitem{zang2020moflow}
C.~Zang and F.~Wang, ``Moflow: an invertible flow model for generating
  molecular graphs,'' in \emph{Proceedings of the 26th ACM SIGKDD International
  Conference on Knowledge Discovery \& Data Mining}, 2020, pp. 617--626.

\bibitem{luo2021graphdf}
Y.~Luo, K.~Yan, and S.~Ji, ``Graphdf: A discrete flow model for molecular graph
  generation,'' in \emph{International Conference on Machine Learning}.\hskip
  1em plus 0.5em minus 0.4em\relax PMLR, 2021, pp. 7192--7203.

\bibitem{creswell2018generative}
A.~Creswell, T.~White, V.~Dumoulin, K.~Arulkumaran, B.~Sengupta, and A.~A.
  Bharath, ``Generative adversarial networks: An overview,'' \emph{IEEE signal
  processing magazine}, vol.~35, no.~1, pp. 53--65, 2018.

\bibitem{mirza2014conditional}
M.~Mirza and S.~Osindero, ``Conditional generative adversarial nets,''
  \emph{arXiv preprint arXiv:1411.1784}, 2014.

\bibitem{radford2015unsupervised}
A.~Radford, L.~Metz, and S.~Chintala, ``Unsupervised representation learning
  with deep convolutional generative adversarial networks,'' \emph{arXiv
  preprint arXiv:1511.06434}, 2015.

\bibitem{salimans2016improved}
T.~Salimans, I.~Goodfellow, W.~Zaremba, V.~Cheung, A.~Radford, and X.~Chen,
  ``Improved techniques for training gans,'' \emph{Advances in neural
  information processing systems}, 2016.

\bibitem{jin2018learning}
W.~Jin, K.~Yang, R.~Barzilay, and T.~Jaakkola, ``Learning multimodal
  graph-to-graph translation for molecular optimization,'' \emph{arXiv preprint
  arXiv:1812.01070}, 2018.

\bibitem{maziarka2020mol}
{\L}.~Maziarka, A.~Pocha, J.~Kaczmarczyk, K.~Rataj, T.~Danel, and
  M.~Warcho{\l}, ``Mol-cyclegan: a generative model for molecular
  optimization,'' \emph{Journal of Cheminformatics}, vol.~12, no.~1, pp. 1--18,
  2020.

\bibitem{ho2020denoising}
J.~Ho, A.~Jain, and P.~Abbeel, ``Denoising diffusion probabilistic models,''
  \emph{Advances in Neural Information Processing Systems}, vol.~33, pp.
  6840--6851, 2020.

\bibitem{vignac2022digress}
C.~Vignac, I.~Krawczuk, A.~Siraudin, B.~Wang, V.~Cevher, and P.~Frossard,
  ``Digress: Discrete denoising diffusion for graph generation,'' \emph{arXiv
  preprint arXiv:2209.14734}, 2022.

\bibitem{haefeli2022diffusion}
K.~K. Haefeli, K.~Martinkus, N.~Perraudin, and R.~Wattenhofer, ``Diffusion
  models for graphs benefit from discrete state spaces,'' \emph{arXiv preprint
  arXiv:2210.01549}, 2022.

\bibitem{anand2022protein}
N.~Anand and T.~Achim, ``Protein structure and sequence generation with
  equivariant denoising diffusion probabilistic models,'' \emph{arXiv preprint
  arXiv:2205.15019}, 2022.

\bibitem{trippe2022diffusion}
B.~L. Trippe, J.~Yim, D.~Tischer, T.~Broderick, D.~Baker, R.~Barzilay, and
  T.~Jaakkola, ``Diffusion probabilistic modeling of protein backbones in 3d
  for the motif-scaffolding problem,'' \emph{arXiv preprint arXiv:2206.04119},
  2022.

\bibitem{luo2022antigen}
S.~Luo, Y.~Su, X.~Peng, S.~Wang, J.~Peng, and J.~Ma, ``Antigen-specific
  antibody design and optimization with diffusion-based generative models,''
  \emph{bioRxiv}, 2022.

\bibitem{niu2020permutation}
C.~Niu, Y.~Song, J.~Song, S.~Zhao, A.~Grover, and S.~Ermon, ``Permutation
  invariant graph generation via score-based generative modeling,'' in
  \emph{AISTATS}.\hskip 1em plus 0.5em minus 0.4em\relax PMLR, 2020, pp.
  4474--4484.

\bibitem{chen2022nvdiff}
X.~Chen, Y.~Li, A.~Zhang, and L.-p. Liu, ``Nvdiff: Graph generation through the
  diffusion of node vectors,'' \emph{arXiv preprint arXiv:2211.10794}, 2022.

\bibitem{huang2022graphgdp}
H.~Huang, L.~Sun, B.~Du, Y.~Fu, and W.~Lv, ``Graphgdp: Generative diffusion
  processes for permutation invariant graph generation,'' \emph{arXiv preprint
  arXiv:2212.01842}, 2022.

\bibitem{jo2022score}
J.~Jo, S.~Lee, and S.~J. Hwang, ``Score-based generative modeling of graphs via
  the system of stochastic differential equations,'' \emph{arXiv preprint
  arXiv:2202.02514}, 2022.

\bibitem{luo2022fast}
T.~Luo, Z.~Mo, and S.~J. Pan, ``Fast graph generative model via spectral
  diffusion,'' \emph{arXiv preprint arXiv:2211.08892}, 2022.

\bibitem{yang2022diffusionsurvey}
L.~Yang, Z.~Zhang, Y.~Song, S.~Hong, R.~Xu, Y.~Zhao, Y.~Shao, W.~Zhang, B.~Cui,
  and M.-H. Yang, ``Diffusion models: A comprehensive survey of methods and
  applications,'' \emph{arXiv preprint arXiv:2209.00796}, 2022.

\bibitem{song2020improved}
Y.~Song and S.~Ermon, ``Improved techniques for training score-based generative
  models,'' \emph{Advances in neural information processing systems}, vol.~33,
  pp. 12\,438--12\,448, 2020.

\bibitem{song2020sliced}
Y.~Song, S.~Garg, J.~Shi, and S.~Ermon, ``Sliced score matching: A scalable
  approach to density and score estimation,'' in \emph{Uncertainty in
  Artificial Intelligence}.\hskip 1em plus 0.5em minus 0.4em\relax PMLR, 2020,
  pp. 574--584.

\bibitem{xu2018powerful}
K.~Xu, W.~Hu, J.~Leskovec, and S.~Jegelka, ``How powerful are graph neural
  networks?'' \emph{arXiv preprint arXiv:1810.00826}, 2018.

\bibitem{song2020score}
Y.~Song, J.~Sohl-Dickstein, D.~P. Kingma, A.~Kumar, S.~Ermon, and B.~Poole,
  ``Score-based generative modeling through stochastic differential
  equations,'' in \emph{International Conference on Learning Representations},
  2020.

\bibitem{li2018learning}
Y.~Li, O.~Vinyals, C.~Dyer, R.~Pascanu, and P.~Battaglia, ``Learning deep
  generative models of graphs,'' \emph{arXiv preprint arXiv:1803.03324}, 2018.

\bibitem{liao2019efficient}
R.~Liao, Y.~Li, Y.~Song, S.~Wang, W.~Hamilton, D.~K. Duvenaud, R.~Urtasun, and
  R.~Zemel, ``Efficient graph generation with graph recurrent attention
  networks,'' \emph{Advances in neural information processing systems},
  vol.~32, 2019.

\bibitem{dai2020scalable}
H.~Dai, A.~Nazi, Y.~Li, B.~Dai, and D.~Schuurmans, ``Scalable deep generative
  modeling for sparse graphs,'' in \emph{International Conference on Machine
  Learning}.\hskip 1em plus 0.5em minus 0.4em\relax PMLR, 2020, pp. 2302--2312.

\bibitem{chen2021order}
X.~Chen, X.~Han, J.~Hu, F.~J. Ruiz, and L.~Liu, ``Order matters: Probabilistic
  modeling of node sequence for graph generation,'' \emph{arXiv preprint
  arXiv:2106.06189}, 2021.

\bibitem{dhariwal2021diffusion}
P.~Dhariwal and A.~Nichol, ``Diffusion models beat gans on image synthesis,''
  vol.~34, 2021, pp. 8780--8794.

\bibitem{alon2020bottleneck}
U.~Alon and E.~Yahav, ``On the bottleneck of graph neural networks and its
  practical implications,'' \emph{arXiv preprint arXiv:2006.05205}, 2020.

\bibitem{jorgensen2018deep}
P.~B. J{\o}rgensen, M.~N. Schmidt, and O.~Winther, ``Deep generative models for
  molecular science,'' \emph{Molecular informatics}, vol.~37, no. 1-2, p.
  1700133, 2018.

\bibitem{xu2022geodiff}
M.~Xu, L.~Yu, Y.~Song, C.~Shi, S.~Ermon, and J.~Tang, ``Geodiff: A geometric
  diffusion model for molecular conformation generation,'' \emph{arXiv preprint
  arXiv:2203.02923}, 2022.

\bibitem{lin2022diffbp}
H.~Lin, Y.~Huang, M.~Liu, X.~Li, S.~Ji, and S.~Z. Li, ``Diffbp: Generative
  diffusion of 3d molecules for target protein binding,'' \emph{arXiv preprint
  arXiv:2211.11214}, 2022.

\bibitem{igashov2022equivariant}
I.~Igashov, H.~St{\"a}rk, C.~Vignac, V.~G. Satorras, P.~Frossard, M.~Welling,
  M.~Bronstein, and B.~Correia, ``Equivariant 3d-conditional diffusion models
  for molecular linker design,'' \emph{arXiv preprint arXiv:2210.05274}, 2022.

\bibitem{hoogeboom2022equivariant}
E.~Hoogeboom, V.~G. Satorras, C.~Vignac, and M.~Welling, ``Equivariant
  diffusion for molecule generation in 3d,'' in \emph{ICML}.\hskip 1em plus
  0.5em minus 0.4em\relax PMLR, 2022, pp. 8867--8887.

\bibitem{satorras2021n}
V.~G. Satorras, E.~Hoogeboom, and M.~Welling, ``E (n) equivariant graph neural
  networks,'' in \emph{International conference on machine learning}.\hskip 1em
  plus 0.5em minus 0.4em\relax PMLR, 2021, pp. 9323--9332.

\bibitem{shi2021learning}
C.~Shi, S.~Luo, M.~Xu, and J.~Tang, ``Learning gradient fields for molecular
  conformation generation,'' 2021, pp. 9558--9568.

\bibitem{walters2020applications}
W.~P. Walters and R.~Barzilay, ``Applications of deep learning in molecule
  generation and molecular property prediction,'' \emph{Accounts of chemical
  research}, vol.~54, no.~2, pp. 263--270, 2020.

\bibitem{atz2021geometric}
K.~Atz, F.~Grisoni, and G.~Schneider, ``Geometric deep learning on molecular
  representations,'' \emph{Nature Machine Intelligence}, vol.~3, no.~12, pp.
  1023--1032, 2021.

\bibitem{zhavoronkov2018artificial}
A.~Zhavoronkov, ``Artificial intelligence for drug discovery, biomarker
  development, and generation of novel chemistry,'' pp. 4311--4313, 2018.

\bibitem{brooks1998markov}
S.~Brooks, ``Markov chain monte carlo method and its application,''
  \emph{Journal of the royal statistical society: series D (the Statistician)},
  vol.~47, no.~1, pp. 69--100, 1998.

\bibitem{xu2021learning}
M.~Xu, S.~Luo, Y.~Bengio, J.~Peng, and J.~Tang, ``Learning neural generative
  dynamics for molecular conformation generation,'' \emph{arXiv preprint
  arXiv:2102.10240}, 2021.

\bibitem{simm2019generative}
G.~N. Simm and J.~M. Hern{\'a}ndez-Lobato, ``A generative model for molecular
  distance geometry,'' \emph{arXiv preprint arXiv:1909.11459}, 2019.

\bibitem{luo2021predicting}
S.~Luo, C.~Shi, M.~Xu, and J.~Tang, ``Predicting molecular conformation via
  dynamic graph score matching,'' vol.~34, 2021, pp. 19\,784--19\,795.

\bibitem{jing2022torsional}
B.~Jing, G.~Corso, J.~Chang, R.~Barzilay, and T.~Jaakkola, ``Torsional
  diffusion for molecular conformer generation,'' \emph{arXiv preprint
  arXiv:2206.01729}, 2022.

\bibitem{elton2019deep}
D.~C. Elton, Z.~Boukouvalas, M.~D. Fuge, and P.~W. Chung, ``Deep learning for
  molecular design—a review of the state of the art,'' \emph{Molecular
  Systems Design \& Engineering}, vol.~4, no.~4, pp. 828--849, 2019.

\bibitem{sanchez2018inverse}
B.~Sanchez-Lengeling and A.~Aspuru-Guzik, ``Inverse molecular design using
  machine learning: Generative models for matter engineering,'' \emph{Science},
  vol. 361, no. 6400, pp. 360--365, 2018.

\bibitem{huang2022mdm}
L.~Huang, H.~Zhang, T.~Xu, and K.-C. Wong, ``Mdm: Molecular diffusion model for
  3d molecule generation,'' \emph{arXiv preprint arXiv:2209.05710}, 2022.

\bibitem{wu2022diffusion}
L.~Wu, C.~Gong, X.~Liu, M.~Ye, and Q.~Liu, ``Diffusion-based molecule
  generation with informative prior bridges,'' \emph{arXiv preprint
  arXiv:2209.00865}, 2022.

\bibitem{anderson2003process}
A.~C. Anderson, ``The process of structure-based drug design,'' \emph{Chemistry
  \& biology}, vol.~10, no.~9, pp. 787--797, 2003.

\bibitem{schneuing2022structure}
A.~Schneuing, Y.~Du, C.~Harris, A.~Jamasb, I.~Igashov, W.~Du, T.~Blundell,
  P.~Li{\'o}, C.~Gomes, M.~Welling \emph{et~al.}, ``Structure-based drug design
  with equivariant diffusion models,'' \emph{arXiv preprint arXiv:2210.13695},
  2022.

\bibitem{bao2022equivariant}
F.~Bao, M.~Zhao, Z.~Hao, P.~Li, C.~Li, and J.~Zhu, ``Equivariant energy-guided
  sde for inverse molecular design,'' \emph{arXiv preprint arXiv:2209.15408},
  2022.

\bibitem{morehead2023geometry}
A.~Morehead and J.~Cheng, ``Geometry-complete diffusion for 3d molecule
  generation,'' \emph{arXiv preprint arXiv:2302.04313}, 2023.

\bibitem{alquraishi2019end}
M.~AlQuraishi, ``End-to-end differentiable learning of protein structure,''
  \emph{Cell systems}, vol.~8, no.~4, pp. 292--301, 2019.

\bibitem{senior2020improved}
A.~W. Senior, R.~Evans, J.~Jumper, J.~Kirkpatrick, L.~Sifre, T.~Green, C.~Qin,
  A.~{\v{Z}}{\'\i}dek, A.~W. Nelson, A.~Bridgland \emph{et~al.}, ``Improved
  protein structure prediction using potentials from deep learning,''
  \emph{Nature}, vol. 577, no. 7792, pp. 706--710, 2020.

\bibitem{jumper2021highly}
J.~Jumper, R.~Evans, A.~Pritzel, T.~Green, M.~Figurnov, O.~Ronneberger,
  K.~Tunyasuvunakool, R.~Bates, A.~{\v{Z}}{\'\i}dek, A.~Potapenko
  \emph{et~al.}, ``Highly accurate protein structure prediction with
  alphafold,'' \emph{Nature}, vol. 596, no. 7873, pp. 583--589, 2021.

\bibitem{yang2020improved}
J.~Yang, I.~Anishchenko, H.~Park, Z.~Peng, S.~Ovchinnikov, and D.~Baker,
  ``Improved protein structure prediction using predicted interresidue
  orientations,'' \emph{Proceedings of the National Academy of Sciences}, vol.
  117, no.~3, pp. 1496--1503, 2020.

\bibitem{mohamed2019transformers}
A.~Mohamed, D.~Okhonko, and L.~Zettlemoyer, ``Transformers with convolutional
  context for asr,'' \emph{arXiv preprint arXiv:1904.11660}, 2019.

\bibitem{rives2021biological}
A.~Rives, J.~Meier, T.~Sercu, S.~Goyal, Z.~Lin, J.~Liu, D.~Guo, M.~Ott, C.~L.
  Zitnick, J.~Ma \emph{et~al.}, ``Biological structure and function emerge from
  scaling unsupervised learning to 250 million protein sequences,''
  \emph{Proceedings of the National Academy of Sciences}, vol. 118, no.~15, p.
  e2016239118, 2021.

\bibitem{wu2022protein}
K.~E. Wu, K.~K. Yang, R.~v.~d. Berg, J.~Y. Zou, A.~X. Lu, and A.~P. Amini,
  ``Protein structure generation via folding diffusion,'' \emph{arXiv preprint
  arXiv:2209.15611}, 2022.

\bibitem{corso2022diffdock}
G.~Corso, H.~St{\"a}rk, B.~Jing, R.~Barzilay, and T.~Jaakkola, ``Diffdock:
  Diffusion steps, twists, and turns for molecular docking,'' \emph{arXiv
  preprint arXiv:2210.01776}, 2022.

\bibitem{zhou2018deep}
J.~Zhou, C.~L. Theesfeld, K.~Yao, K.~M. Chen, A.~K. Wong, and O.~G.
  Troyanskaya, ``Deep learning sequence-based ab initio prediction of variant
  effects on expression and disease risk,'' \emph{Nature genetics}, vol.~50,
  no.~8, pp. 1171--1179, 2018.

\bibitem{north2011new}
B.~North, A.~Lehmann, and R.~L. Dunbrack~Jr, ``A new clustering of antibody cdr
  loop conformations,'' \emph{Journal of molecular biology}, vol. 406, no.~2,
  pp. 228--256, 2011.

\bibitem{lefranc2003imgt}
M.-P. Lefranc, C.~Pommi{\'e}, M.~Ruiz, V.~Giudicelli, E.~Foulquier, L.~Truong,
  V.~Thouvenin-Contet, and G.~Lefranc, ``Imgt unique numbering for
  immunoglobulin and t cell receptor variable domains and ig superfamily v-like
  domains,'' \emph{Developmental \& Comparative Immunology}, vol.~27, no.~1,
  pp. 55--77, 2003.

\bibitem{dunbar2014sabdab}
J.~Dunbar, K.~Krawczyk, J.~Leem, T.~Baker, A.~Fuchs, G.~Georges, J.~Shi, and
  C.~M. Deane, ``Sabdab: the structural antibody database,'' \emph{Nucleic
  acids research}, vol.~42, no.~D1, pp. D1140--D1146, 2014.

\bibitem{mobley2009binding}
D.~L. Mobley and K.~A. Dill, ``Binding of small-molecule ligands to
  proteins:“what you see” is not always “what you get”,''
  \emph{Structure}, vol.~17, no.~4, pp. 489--498, 2009.

\bibitem{qiao2022dynamic}
Z.~Qiao, W.~Nie, A.~Vahdat, T.~F. Miller~III, and A.~Anandkumar,
  ``Dynamic-backbone protein-ligand structure prediction with multiscale
  generative diffusion models,'' \emph{arXiv preprint arXiv:2209.15171}, 2022.

\bibitem{glusker1996crystal}
J.~P. Glusker, M.~Lewis, and M.~Rossi, \emph{Crystal structure analysis for
  chemists and biologists}.\hskip 1em plus 0.5em minus 0.4em\relax John Wiley
  \& Sons, 1996, vol.~16.

\bibitem{xie2021crystal}
T.~Xie, X.~Fu, O.-E. Ganea, R.~Barzilay, and T.~Jaakkola, ``Crystal diffusion
  variational autoencoder for periodic material generation,'' \emph{arXiv
  preprint arXiv:2110.06197}, 2021.

\bibitem{song2019generative}
Y.~Song and S.~Ermon, ``Generative modeling by estimating gradients of the data
  distribution,'' vol.~32, 2019.

\bibitem{sutherland2016generative}
D.~J. Sutherland, H.-Y. Tung, H.~Strathmann, S.~De, A.~Ramdas, A.~Smola, and
  A.~Gretton, ``Generative models and model criticism via optimized maximum
  mean discrepancy,'' \emph{arXiv preprint arXiv:1611.04488}, 2016.

\bibitem{theis2015note}
L.~Theis, A.~v.~d. Oord, and M.~Bethge, ``A note on the evaluation of
  generative models,'' \emph{arXiv preprint arXiv:1511.01844}, 2015.

\bibitem{sajjadi2018assessing}
M.~S. Sajjadi, O.~Bachem, M.~Lucic, O.~Bousquet, and S.~Gelly, ``Assessing
  generative models via precision and recall,'' \emph{Advances in neural
  information processing systems}, vol.~31, 2018.

\bibitem{zhou2020graph}
J.~Zhou, G.~Cui, S.~Hu, Z.~Zhang, C.~Yang, Z.~Liu, L.~Wang, C.~Li, and M.~Sun,
  ``Graph neural networks: A review of methods and applications,'' \emph{AI
  Open}, vol.~1, pp. 57--81, 2020.

\bibitem{zhang2019graph}
S.~Zhang, H.~Tong, J.~Xu, and R.~Maciejewski, ``Graph convolutional networks: a
  comprehensive review,'' \emph{Computational Social Networks}, vol.~6, no.~1,
  pp. 1--23, 2019.

\bibitem{clauset2004finding}
A.~Clauset, M.~E. Newman, and C.~Moore, ``Finding community structure in very
  large networks,'' \emph{Physical review E}, vol.~70, no.~6, p. 066111, 2004.

\bibitem{karrer2011stochastic}
B.~Karrer and M.~E. Newman, ``Stochastic blockmodels and community structure in
  networks,'' \emph{Physical review E}, vol.~83, no.~1, p. 016107, 2011.

\bibitem{xu2018graph2seq}
K.~Xu, L.~Wu, Z.~Wang, Y.~Feng, M.~Witbrock, and V.~Sheinin, ``Graph2seq: Graph
  to sequence learning with attention-based neural networks,'' \emph{arXiv
  preprint arXiv:1804.00823}, 2018.

\bibitem{wang2016structural}
D.~Wang, P.~Cui, and W.~Zhu, ``Structural deep network embedding,'' in
  \emph{Proceedings of the 22nd ACM SIGKDD international conference on
  Knowledge discovery and data mining}, 2016, pp. 1225--1234.

\bibitem{gomez2018automatic}
R.~G{\'o}mez-Bombarelli, J.~N. Wei, D.~Duvenaud, J.~M. Hern{\'a}ndez-Lobato,
  B.~S{\'a}nchez-Lengeling, D.~Sheberla, J.~Aguilera-Iparraguirre, T.~D.
  Hirzel, R.~P. Adams, and A.~Aspuru-Guzik, ``Automatic chemical design using a
  data-driven continuous representation of molecules,'' \emph{ACS central
  science}, vol.~4, no.~2, pp. 268--276, 2018.

\bibitem{velivckovic2017graph}
P.~Veli{\v{c}}kovi{\'c}, G.~Cucurull, A.~Casanova, A.~Romero, P.~Lio, and
  Y.~Bengio, ``Graph attention networks,'' \emph{arXiv preprint
  arXiv:1710.10903}, 2017.

\bibitem{velickovic2019deep}
P.~Velickovic, W.~Fedus, W.~L. Hamilton, P.~Li{\`o}, Y.~Bengio, and R.~D.
  Hjelm, ``Deep graph infomax.'' \emph{ICLR (Poster)}, vol.~2, no.~3, p.~4,
  2019.

\bibitem{perozzi2014deepwalk}
B.~Perozzi, R.~Al-Rfou, and S.~Skiena, ``Deepwalk: Online learning of social
  representations,'' in \emph{Proceedings of the 20th ACM SIGKDD international
  conference on Knowledge discovery and data mining}, 2014, pp. 701--710.

\end{thebibliography}

\ifCLASSOPTIONcaptionsoff
  \newpage
\fi

\end{document}